\documentclass[11pt,a4paper]{article}
\usepackage[hyperref]{acl2018}

\usepackage[T1]{fontenc}
\usepackage[utf8]{inputenc}
\usepackage{times}
\usepackage{latexsym}
\usepackage{url}
\usepackage{amsmath, amsthm, amssymb}
\usepackage{algorithm}
\usepackage{algpseudocode}

\usepackage{graphicx, subcaption}
\usepackage{array, booktabs, multirow}
\usepackage{enumitem}
\usepackage{nameref}  
\usepackage{xspace}  
\usepackage{longtable} 
\graphicspath{{figures/}}

\makeatletter
\DeclareRobustCommand\onedot{\futurelet\@let@token\@onedot}
\def\@onedot{\ifx\@let@token.\else.\null\fi\xspace}

\def\eg{\emph{e.g}\onedot} 
\def\ie{\emph{i.e}\onedot}

\def\wrt{w.r.t\onedot} 

\makeatother
\newcommand{\cid}{\textsc{CIDEr}~}
\newcommand{\coco}{\textsc{MS-COCO}~}
\newcommand{\bleu}{\textsc{BLEU}}
\newcommand{\bleuf}{\textsc{BLEU-4}~}

\def\bea{\begin{eqnarray}}
\def\eea{\end{eqnarray}}

\def\fig#1{Figure~\ref{fig:#1}}
\def\tab#1{Table~\ref{tab:#1}}
\def\sect#1{Section~\ref{sec:#1}}

\def\Eq#1{Eq.~(\ref{eq:#1})}
\def\app#1{Appendix~\ref{appendix:#1}}
\newcommand{\parref}[1]{Section~\ref{para:#1}, \nameref{para:#1}}

\newcommand{\ens}{\raisebox{.6\height}{\scalebox{1}{+}}}
\newcommand{\rl}{\raisebox{.4\height}{\scalebox{1}{$\dagger$}}}
\newcommand{\add}{\raisebox{.4\height}{\scalebox{1}{$\circ$}}}

\newcommand{\E}[2]{\mathbb{E}_{#1}\!\left[#2\right]}
\DeclareMathOperator\freq{freq}

\newcommand{\ham}{d}
\newcommand{\rt}{r}
\newcommand{\q}{q}
\newcommand{\V}{\mathcal V}
\newcommand{\R}{\mathbb{R}}
\newcommand{\sml}[2]{\ell_{\text{#1}}^{#2}}
\newcommand{\spl}[1]{\ell_{\text{#1}}}
\newcommand{\ml}{\spl{MLE}}
\newcommand{\gt}[1]{{y_{#1}^*}}
\newcommand{\cond}{\gt{},x}
\newcommand{\p}{p_\theta}

\newcolumntype{H}{>{\setbox0=\hbox\bgroup}c<{\egroup}@{}}

\newcommand{\compresslist}{ 
\setlength{\itemsep}{1pt}
\setlength{\parskip}{1pt}
\setlength{\parsep}{0pt}
}

\aclfinalcopy 
\def\aclpaperid{1547} 

\title{Token-level and sequence-level loss smoothing for RNN language models}

\author{Maha Elbayad\textsuperscript{1,2} \hspace{20pt}
        Laurent Besacier\textsuperscript{1} \hspace{20pt}
        Jakob Verbeek\textsuperscript{2} \\
        Univ.\ Grenoble Alpes, CNRS, Grenoble INP, Inria, LIG, LJK, F-38000 Grenoble France\\
        \textsuperscript{1} \tt{firstname.lastname@univ-grenoble-alpes.fr}\\
        \textsuperscript{2} \tt{firstname.lastname@inria.fr}
         }

\date{}

\begin{document}
\maketitle

\begin{abstract}
Despite the effectiveness of recurrent neural network language models, their 
maximum likelihood estimation suffers from two limitations. It treats all sentences that do not match the ground truth as equally poor, ignoring the structure of the output space. 
Second, it suffers from ``exposure bias'': during training tokens are predicted given ground-truth sequences, while at test time prediction is conditioned on generated output sequences.
To overcome these limitations 
we build upon the recent reward augmented maximum likelihood  approach \ie sequence-level smoothing that encourages the model to predict sentences close to the ground truth according to a given performance metric.
We extend this approach to token-level loss smoothing, and propose improvements to the sequence-level smoothing approach. 
Our experiments on two different tasks, image captioning and machine translation,  show that token-level and sequence-level loss smoothing are complementary, and significantly  improve results.
\end{abstract}

\section{Introduction}

Recurrent neural networks (RNNs) have recently proven to be very effective sequence modeling tools, and are now state of the art for tasks such as machine translation \citep{cho14emnlp,sutskever14nips, bahdanau15iclr}, image captioning \citep{kiros14icml,vinyals15cvpr,anderson17arxiv} and automatic speech recognition \citep{chorowski15nips,chiu2017}.

The basic principle of RNNs is to iteratively compute a vectorial sequence representation, by applying at each time-step the same trainable function to compute the new network state from the previous state and the last symbol in the sequence.   
These models are typically trained by maximizing the likelihood of the target sentence  given an encoded source (text, image, speech). 

Maximum likelihood estimation (MLE), however, has two main limitations. First, the training signal only differentiates the ground-truth target output from all other outputs.  It treats all other output sequences as equally incorrect, regardless of their semantic proximity from the ground-truth target.
While such a ``zero-one'' loss is probably acceptable for coarse grained  classification of images, \eg across a limited number of basic object categories \citep{everingham10ijcv} it becomes problematic as the output space becomes larger and some of its elements become semantically similar to each other.
This is in particular the case for tasks that involve natural language generation (captioning, translation, speech recognition) where the number of possible outputs is practically unbounded. 
For natural language generation tasks, evaluation measures typically do take into account structural similarity, \eg based on n-grams, but such structural information is not reflected in the MLE criterion.
The second limitation of MLE is that training is based on predicting the next token given the input and preceding \emph{ground-truth} output tokens, while at test time the model predicts conditioned on the input and the so-far \emph{generated} output sequence. Given the exponentially large output space of natural language sentences, it is not obvious that the learned RNNs generalize well beyond the relatively sparse distribution of ground-truth sequences used during MLE optimization. 
This phenomenon is known as ``exposure bias''~\citep{ranzato16iclr, bengio15nips}. 

MLE minimizes the KL divergence between a target Dirac distribution on the ground-truth sentence(s) and the model's distribution.
In this paper, we build upon the ``loss smoothing'' approach by \citet{norouzi16nips}, which smooths the Dirac target distribution over  similar sentences, increasing the support of the training data in the output space. We make the following main contributions:
\begin{itemize}[noitemsep,nolistsep]
\item We propose a token-level loss smoothing approach, using word-embeddings, to achieve smoothing among semantically similar terms, and we introduce a special procedure to promote rare tokens.
\item For sequence-level smoothing, we propose to use restricted token replacement vocabularies, and a ``lazy evaluation'' method that significantly speeds up training. 
\item 
We experimentally validate our approach on the MSCOCO image captioning task and the WMT'14 English to French machine translation task, showing that on both tasks combining token-level and sequence-level loss smoothing improves results significantly over maximum likelihood baselines.
\end{itemize}

In the remainder of the paper, we review the existing methods to improve RNN training in \sect{related}. Then, we present  our token-level and sequence-level approaches in \sect{contrib}.  Experimental evaluation  results based on image captioning and machine translation tasks are laid out in \sect{exp}.

\section{Related work}\label{sec:related}

Previous work aiming to improve the generalization performance of  RNNs can be roughly divided into three categories: those  based on regularization, data augmentation, and alternatives to maximum likelihood estimation.

Regularization techniques are used to increase the smoothness of the function learned by the network, \eg by imposing an $\ell_2$ penalty on the network weights, also known as ``weight decay''. 
More recent approaches mask network activations during training, as in dropout \citep{srivastava14jmlr} and its variants adapted to recurrent models \citep{rnn-dropout-a, krueger17iclr}. Instead of masking, batch-normalization \citep{ioffe15icml} rescales the network activations to avoid saturating the network's non-linearities. 
Instead of regularizing the network parameters or activations, it is also possible to directly regularize based on the entropy of the output distribution 
\citep{pereyra17iclr}.

Data augmentation techniques improve the robustness of the learned models  by applying transformations that might be encountered at test time to the training data.
In computer vision, this is common practice, and implemented by, \eg, scaling, cropping, and rotating training images \citep{lecun98ieee,krizhevsky12nips,paulin14cvpr}.
In natural language processing, examples of data augmentation include input noising by randomly dropping some input tokens \citep{iyyer15acl,bowman15CoNLL,kumar16icml}, 
and randomly replacing words with substitutes sampled from the model \citep{bengio15nips}. 
\citet{xie17iclr} introduced data augmentation schemes for RNN language models that leverage n-gram statistics in order to mimic Kneser-Ney smoothing of n-grams models. 
In the context of machine translation, \citet{fadaee17acl} modify sentences by replacing words with rare ones when this is  plausible according to a pre-trained language model, and substitutes its equivalent in the target sentence using automatic word alignments. This approach, however, relies 
on the availability of additional monolingual data for language model training.

The \emph{de facto} standard way to train RNN language models is maximum likelihood estimation (MLE) \citep{cho14emnlp, sutskever14nips, bahdanau15iclr}. 
The sequential factorization of the sequence likelihood generates an additive structure in the loss, with one term corresponding to the prediction of each output token given the input and the preceding ground-truth output tokens. 
In order to directly optimize for sequence-level structured loss functions, such as measures based on n-grams like \textsc{Bleu} or \textsc{Cider},
\citet{ranzato16iclr} use reinforcement learning techniques that optimize the expectation of a sequence-level reward.
In order to avoid early convergence to poor local optima, they pre-train the model using MLE.

\citet{leblond18iclr} build on the learning to search approach to structured prediction \citep{daume09ml,chang15icml} and adapts it to RNN training. The model generates candidate sequences at each time-step using all possible tokens, and scores these at sequence-level to derive a training signal for each time step. 
This leads to an approach that is structurally close to MLE, but computationally expensive.
\citet{norouzi16nips} introduce a reward augmented maximum likelihood (RAML) approach, that incorporates a notion of sequence-level reward without facing the difficulties of reinforcement learning. They define a target distribution  over output sentences using a soft-max over the reward over all possible outputs. Then, they minimize the KL divergence between the target distribution and the model's output distribution. 
Training with a general reward distribution is similar to MLE training, except that we use multiple sentences sampled from the target distribution instead of only the ground-truth sentences.

In our work, we build upon the work of \citet{norouzi16nips} by proposing improvements to sequence-level smoothing, and extending it to token-level smoothing. 
Our token-level smoothing approach is related to the label smoothing approach of \citet{szegedy16cvpr}  for image classification. Instead of maximizing the  probability of the correct class, they train the model to predict the correct class with a large probability and all other classes with a small uniform probability. This regularizes the model by preventing over-confident predictions.
In natural language generation with large vocabularies, preventing such ``narrow'' over-confident distributions is imperative, since for many tokens there are nearly interchangeable alternatives.

\section{Loss smoothing for RNN training}\label{sec:contrib}

We briefly recall standard recurrent neural network training, before presenting sequence-level and token-level loss smoothing below.

\subsection{Maximum likelihood RNN training}\label{sec:prelim}

We are interested in modeling the conditional probability of a sequence $y=(y_1,\dots, y_T)$ given a conditioning observation $x$,
\begin{equation}
\p(y|x) = \prod_{t=1}^T \p(y_t |x, y_{<t}),
\end{equation}
where  $y_{<t}=(y_1,\dots, y_{t-1})$, the model parameters are given by $\theta$, and $x$ is a source sentence or an image in the contexts of machine translation and image captioning, respectively. 

In a recurrent neural network, the sequence $y$ is predicted based on 
a sequence of states $h_t$,
\begin{equation}\label{eq:classifier}
    p_\theta(y_t|x,y_{<t}) = p_\theta(y_t|h_t),
\end{equation}
where the RNN state is computed recursively as
\begin{equation}
h_t = \begin{cases}
      f_\theta(h_{t-1}, y_{t-1},x) & \text{for } t \in \{1,..T\},\label{eq:rnn} \\
      g_\theta(x) & \text{for } t=0.
\end{cases}
\end{equation}
The input is encoded by  $g_\theta$ and used to initialize the state sequence, and $f_\theta$ is a non-linear function that updates the state given the previous state $h_{t-1}$, the last output token $y_{t-1}$, and possibly the input $x$. 
The state update function can take different forms, the ones including gating mechanisms such as LSTMs~\citep{hochreiter97neco} and GRUs~\citep{chung14dlw} are particularly effective to model long sequences.

In standard teacher-forced training, the hidden states will be computed by forwarding the ground truth sequence $\gt{}$ \ie in \Eq{rnn}, the RNN has access to the true previous token $\gt{t-1}$. In this case we will note the hidden states $h_t^*$.

Given a ground-truth target sequence $y^*$, maximum likelihood estimation (MLE) of the network parameters $\theta$ amounts 
to minimizing the loss
\begin{align}
\ml(y^*,x) & =  -\ln \p(y^* |x) \\
           & =  -\sum_{t=1}^T \ln \p(y^*_t |h_t^*). 
\end{align}
The loss can equivalently be expressed as the  KL-divergence between a Dirac centered on the target output (with $\delta_a(x) = 1$ at $x=a$ and $0$ otherwise) and the model distribution, either at the sequence-level or at the token-level:
\begin{align}
\ml(y^*,x) & =   D_\text{KL}\big(\delta_{y^*} || \p(y|x)\big) \label{eq:klMLE} \\
           & =  \sum_{t=1}^T D_\text{KL}\big(\delta_{y_t^*} || \p(y_t|h_t^*)\big).\label{eq:dirac_tok}
\end{align}
Loss smoothing approaches considered in this paper consist in replacing the Dirac on the ground-truth sequence with distributions with larger support. 
These distributions can be designed in such a manner that they reflect which deviations from ground-truth predictions are preferred over others.

\subsection{Sequence-level loss smoothing}\label{sec:seq}

The reward augmented maximum likelihood approach of \citet{norouzi16nips} consists in replacing the sequence-level Dirac $\delta_{y^*}$ in \Eq{klMLE} with a distribution 
\begin{equation}\label{eq:rewardDistr}
r(y|y^*) \propto \exp r(y,y^*) / \tau,
\end{equation}
where $r(y,y^*)$ is a ``reward'' function that measures the quality of sequence $y$ \wrt $y^*$, \eg metrics used for evaluation of natural language processing tasks can be used, such as \bleu~\citep{papineni02acl} or \cid~\citep{vedantam15cvpr}. 
The temperature parameter $\tau$ controls the concentration of the distribution around $y^*$.
When $m>1$ ground-truth sequences are paired with the same input $x$, the reward function can be adapted to fit this setting and be defined as $r(y,\{y^{*{(1)}}, \dots ,y^{*(m)}\})$.
The sequence-level smoothed loss function is then given by 
\begin{align}
\ell_\text{Seq}(y^*,x) & = D_\text{KL}\big(r(y|y^*)||\p(y|x)\big) \nonumber\\
                       &  =  H( r(y|y^*)) - \E{r}{\ln \p(y|x)},\label{eq:seq2}
\end{align}
where the entropy term $H(r(y|y^*))$ does not depend on the model parameters $\theta$.

In general, expectation in \Eq{seq2} is intractable due to the exponentially large output space, and replaced with a Monte-Carlo approximation:
\begin{equation}\label{eq:mc}
    \E{r}{-\ln \p(y|x)}  \approx  -\sum_{l=1}^L \ln \p(y^l|x).
\end{equation}

\paragraph{Stratified sampling.}\label{para:strat}
\citet{norouzi16nips} show that when using the Hamming or edit distance as a reward, we can sample directly from $r(y|y^*)$ using a stratified sampling approach. In this case sampling proceeds in three stages. 
(i)   Sample a distance $d$ from $\{0,\ldots, T\}$ from a prior distribution on $d$.
(ii)   Uniformly select $d$ positions in the sequence to be modified.
(iii)   Sample the $d$ substitutions uniformly from the token vocabulary.

Details on the construction of the prior distribution on $d$ for a reward based on the Hamming distance can be found in \app{strat}.

\paragraph{Importance sampling.}\label{para:importance}
For a reward based on BLEU or \cid, we cannot directly sample from $r(y|y^*)$ since the normalizing constant, or ``partition function'', of the distribution is intractable to compute.
In this case we can resort to importance sampling.
We first sample $L$ sequences $y^l$ from a tractable proposal distribution $\q(y|y^*)$.
We then compute the importance weights
\begin{equation}
\omega_l \approx \frac{ \rt(y^l | \gt{}) / \q(y^l | \gt{})}{\sum_{k=1}^L \rt(y^k | \gt{}) / \q(y^k| \gt{})}, 
\end{equation}
where $r(y^k| \gt{})$ is the un-normalized reward distribution in \Eq{rewardDistr}.
We  finally approximate the expectation by reweighing  the samples  in the Monte Carlo approximation as
\begin{equation}\label{eq:importance}
    \E{r}{-\ln \p(y|x)}  \approx  -\sum_{l=1}^L \omega_l \ln \p(y^l|x).
\end{equation}

In our experiments we use a proposal distribution based on the Hamming distance, which allows for tractable stratified sampling, and generates sentences that do not stray away from the ground truth.

We  propose two modifications to the sequence-level loss smoothing of \citet{norouzi16nips}: sampling to a restricted vocabulary (described in the following paragraph) and lazy sequence-level smoothing (described in section \ref{sec:combining}).

\paragraph{Restricted vocabulary sampling.}\label{para:subset}
In the stratified sampling method for Hamming and edit distance rewards, instead of drawing from the large vocabulary $\V$, containing typically in the order of $10^4$ words or more, we can restrict ourselves to a smaller subset $\V_{sub}$ more adapted to our task. We considered three different possibilities for $\V_{sub}$.

   $\V$ : the full vocabulary from which we sample uniformly (default), or draw from our token-level smoothing distribution defined below in \Eq{rtok}. 
    
    $\V_{refs}$: uniformly sample from the set of tokens that appear in the ground-truth sentence(s) associated with the current input. 

$\V_{batch}$: uniformly sample from the tokens that appear in the ground-truth sentences across all inputs that appear in a given training mini-batch.

Uniformly sampling from $\V_{batch}$ has the effect of boosting the frequencies of words that appear in many reference sentences, 
and thus approximates to some extent sampling substitutions from the uni-gram statistics of the training set.

\subsection{Token-level loss smoothing}\label{sec:tok}

While the sequence-level smoothing can be directly based on performance measures of interest such as BLEU or CIDEr, the support of the smoothed distribution is limited to the number of samples drawn during training. 
We propose smoothing  the token-level Diracs $\delta_{y_t^*}$ in \Eq{dirac_tok} to increase its support to similar tokens. 
Since we apply smoothing to each of the tokens independently, this approach implicitly increases the support to an exponential number of sequences, unlike the sequence-level smoothing approach. 
This comes at the price, however, of a naive token-level  independence assumption in the smoothing.

We define the smoothed token-level distribution, similar as the sequence-level one, as a soft-max over a token-level ``reward'' function,
\begin{equation}\label{eq:rtok}
    \rt(y_t| \gt t) \propto  \exp r(y_t, \gt t) / \tau,
\end{equation}
where $\tau$ is again a temperature parameter. 
As a  token-level reward $r(y_t, \gt t)$ we use the cosine similarity between 
$y_t$ and $\gt t$ in a semantic word-embedding space. 
In our experiments we use  GloVe \citep{pennington14emnlp}; preliminary experiments with  word2vec \citep{mikolov13iclr} yielded somewhat worse results.

\paragraph{Promoting rare tokens.}
We can further improve the token-level smoothing by promoting rare tokens. 
To do so, we penalize frequent tokens when smoothing over the vocabulary, by subtracting $\beta\freq(y_t)$ from the reward, where  $\freq(\cdot)$ denotes the term frequency and $\beta$ is a non-negative weight. 
This modification encourages frequent tokens into considering the rare ones. 
We experimentally found that it is also beneficial for rare tokens to boost frequent ones, as they tend to have mostly rare tokens as neighbors in the word-embedding space. 
With this in mind, we define a new token-level reward as:
\begin{align}
    r^{\freq}(y_t,\gt t) & = r(y_t, \gt t) \\
                        & - \beta \min\left(\frac{\freq(y_t)}{\freq(\gt t)}, \frac{\freq(\gt t)}{\freq(y_t)}\right), \nonumber
\end{align}
where the penalty term is strongest if both tokens have similar frequencies.

\subsection{Combining losses}\label{sec:combining}

In both loss smoothing methods presented above, the temperature parameter $\tau$ controls the concentration of the distribution. 
As  $\tau$ gets smaller the distribution peaks around the ground-truth, while for large $\tau$ the uniform distribution is approached.
We can, however, not separately control the spread of the distribution and the mass reserved for the ground-truth output. 
We therefore introduce a second parameter $\alpha\in [0,1]$ to interpolate between the Dirac on the ground-truth and the smooth distribution. 
Using $\bar\alpha=1-\alpha$, the sequence-level and token-level loss functions are then defined as
\begin{align}
    \sml{Seq}{\alpha}(\cond) & =  \alpha \spl{Seq}(\cond) + \bar\alpha  \ml(\cond)\\
                             & = \alpha \E{r}{\ml(y, x)}+ \bar\alpha  \ml(\cond)\nonumber\\
    \sml{Tok}{\alpha}(\cond) &  =  \alpha \spl{Tok}(\cond) + \bar\alpha \ml(\cond)
\end{align}

To benefit from both sequence-level and token-level loss smoothing, 
we also combine them by applying token-level smoothing to the different sequences sampled for the sequence-level smoothing. 
We introduce  two mixing parameters $\alpha_1$ and $\alpha_2$.
The first controls to what extent sequence-level smoothing is used, while the second controls to what extent token-level smoothing is used. 
The combined loss is defined as 
\begin{align}
    \sml{Seq, Tok}{\alpha_1, \alpha_2}(\cond, \rt) & = \alpha_1  \E{\rt}{\spl{Tok}(y, x)} + \bar\alpha_1 \spl{Tok}(\cond)\nonumber\\
                                                   & = \alpha_1 \E{\rt}{ \alpha_2\spl{Tok}(y, x) + \bar\alpha_2\ml(y, x)}\nonumber\\
                                                   & + \bar\alpha_1 (\alpha_2\spl{Tok}(\cond)+\bar\alpha_2\ml(\cond)).
\end{align}

In our experiments, we use held out validation data to set mixing and temperature parameters. 

\paragraph{Lazy sequence smoothing.}\label{para:lazy}
\begin{algorithm}
    \caption{Sequence-level smoothing algorithm}
		\label{alg:seq}
		\begin{algorithmic}
        \small
        \Require $x, \gt{}$
        \Ensure $\sml{seq}{\alpha}(x,\gt{})$
        \State Encode $x$ to initialize the RNN
        \State Forward $\gt{}$ in the RNN to compute the hidden states $h_t^*$
        \State Compute the MLE loss $\ml(\gt{},x)$
		\For{$l\in \{1,\ldots, L\}$}
            \State Sample $y^l \sim r(\dot|\gt{})$\;
            \If{Lazy} 
                \State Compute $\ell(y^l, x)=-\sum_t \log\p(y_t^l|h_t^*)$
             \Else
                \State Forward $y^l$ in the RNN to get its hidden states $h^l_t$
                \State Compute $\ell(y^l, x)=\ml(y^l,x)$
            \EndIf
        \EndFor
        \State $\sml{Seq}{\alpha}(x,\gt{})= \bar\alpha \ml(\gt{},x) + \frac{\alpha}{L}\sum_l \ell(y^l, x)$
       \end{algorithmic}
\end{algorithm}

    Although sequence-level smoothing is computationally efficient compared to reinforcement learning approaches \citep{ranzato16iclr, rennie17cvpr}, it is slower compared to MLE. In particular, we need to forward each of the samples $y^l$  through the RNN in teacher-forcing mode so as to compute its hidden states $h^l_t$, which are used to compute the sequence MLE loss as 
    \begin{equation}
\ml(y^l,x) = -\sum_{t=1}^T\ln \p(y^l_t | h_t^l).
        \label{eq:sample_loss}
    \end{equation}
    To speed up training, and since we already forward the ground truth sequence in the RNN to evaluate the MLE part of $\sml{Seq}{\alpha}(\gt{},x)$, we propose to use the same hidden states $h_t^*$ to compute both the MLE and the sequence-level smoothed loss. In this case:
     \begin{equation}
         \ell_{\text{lazy}}(y^l, x) = - \sum_{t=1}^T\ln \p(y^l_t | h_t^*)
    \end{equation}
In this manner, we only have a single instead of $L+1$ forwards-passes in the RNN.
 We provide the  pseudo-code for training in Algorithm~\ref{alg:seq}.
\begin{table*}[t]
\centering
\resizebox*{!}{.38\textheight}{
\begin{tabular}{HHp{3.5cm}llHHccHHHc}
\cmidrule[.8pt]{6-13} 
\multicolumn{5}{c}{} & \multicolumn{8}{c}{\bf Without attention}\\
\midrule
Model & Init & Loss & Reward & $\V_{sub}$ & $Bleu4_ph1$ & $CIDEr_ph1$ &  \bleu{-1} & \bleu{-4}  & ROUGE-L & SPICE & METEOR & \cid\\
\midrule
Show \& Tell & random & MLE &&& 28.34 & 88.30 & 70.63 & 30.14 & 52.37 & 17.66 & 24.75 & 93.59\\
Show \& Tell & random & MLE + $\gamma H$  &&& 28.59 & 88.17 & 70.79 & 30.29 & 52.35 & 17.76 & 24.89 & 93.61\\
\midrule
\midrule
Show \& Tell & random & Tok & Glove sim && 30.07 & 92.36 & 71.94 & 31.27 & 52.75 & 17.62 & 24.69 & 95.79\\
Show \& Tell & random & Tok & Glove sim $r^{\freq}$ && 30.09 & 92.73 & 72.39 & 31.76 & 53.35 & 17.90 & 25.00 & 97.47\\
\midrule
\midrule
Show \& Tell & random & Seq & Hamming & $\V$ & 29.08 & 90.08 & 71.76 & 31.16 & 52.99 & 18.08 & 25.19 & 96.37\\
Show \& Tell & random & Seq & Hamming & $\V_{batch}$ & 29.44 & 90.79 & 71.46 & 31.15 & 52.93 & 18.03 & 25.13 & \bf 96.53\\
Show \& Tell & random & Seq & Hamming & $\V_{refs}$ & 29.88 & 91.66 & \bf 71.80 & \bf 31.63 & 53.00 & 17.89 & 25.05 & 96.22\\
\midrule
Show \& Tell & random & Seq, lazy & Hamming & $\V$ & 29.11 & 89.60 & 70.81 & 30.43 & 52.24 & 17.63 & 24.74 & 94.26\\
Show \& Tell & random & Seq, lazy & Hamming & $\V_{batch}$ & 29.14 & 90.51 & 71.85 & 31.13 & 52.96 & 18.00 & 25.20 & \bf 96.65\\
Show \& Tell & random & Seq, lazy & Hamming & $\V_{refs}$ & 29.67 & 90.79 & \bf 71.96 & \bf 31.23 & 52.90 & 17.99 & 24.99 & 95.34\\
\midrule
Show \& Tell & random & Seq & \cid & $\V$ & 29.29 & 91.01 & 71.05 & 30.46 & 52.38 & 17.69 & 24.86 & 94.40\\
Show \& Tell & random & Seq & \cid & $\V_{batch}$ & 29.46 & 91.07 & 71.51 & 31.17 & 52.80 & 17.89 & 25.02 & 95.78\\
Show \& Tell & random & Seq & \cid & $\V_{refs}$ & 30.04 & 91.99 & \bf 71.93 & \bf 31.41 & 53.16 & 17.93 & 25.08 & \bf 96.81\\
\midrule
Show \& Tell & random & Seq, lazy & \cid & $\V$ & 29.24 & 90.30 & 71.43 & \bf 31.18 & 52.88 & 18.00 & 25.17 & \bf 96.32\\
Show \& Tell & random & Seq, lazy & \cid & $\V_{batch}$ & 29.29 & 90.81 & 71.47 & 31.00 & 52.94 & 17.90 & 25.07 & 95.56\\
Show \& Tell & random & Seq, lazy & \cid & $\V_{refs}$ & 29.63 & 90.75 & \bf 71.82 & 31.06 & 52.89 & 17.87 & 24.91 & 95.66\\
\midrule
\midrule
Show \& Tell & random & Tok-Seq & Hamming & $\V$ & 29.38 & 92.86 & 70.79 & 30.43 & 0 & 0 & 0 & 96.34\\
Show \& Tell & random & Tok-Seq & Hamming & $\V_{batch}$ & 29.90 & 91.15 & 72.28 & 31.65 & 53.19 & 17.72 & 24.90 & 96.73\\
Show \& Tell & random & Tok-Seq & Hamming & $\V_{refs}$ & 30.48 & 92.82 & 72.69 & 32.30 & 53.52 & 17.87 & 25.01 & 98.01\\
Show \& Tell & random & Tok-Seq & \cid & $\V$ & 29.39 & 92.25 & 70.80 & 30.55 & 0 & 0 & 0 & 96.89\\
Show \& Tell & random & Tok-Seq & \cid & $\V_{batch}$ & 30.24 & 92.52 & 72.13 & 31.71 & 53.06 & 17.70 & 24.79 & 96.92\\
Show \& Tell & random & Tok-Seq & \cid & $\V_{refs}$ & 30.31 & 93.16 & \bf 73.08 & \bf 32.82 & 53.86 & 18.22 & 25.18 &\bf 99.92\\
\bottomrule
\end{tabular}
}
\resizebox*{!}{.38\textheight}{
\begin{tabular}{HHHHHHHccHHHc}
\cmidrule[.8pt]{1-13} 
\multicolumn{13}{c}{\bf With attention}\\
\midrule
Model & Init & Loss & Reward & $\V_{sub}$ & $Bleu4_ph1$ & $CIDEr_ph1$ & \bleu{-1} & \bleu{-4} & ROUGE-L & SPICE & METEOR & \cid\\
\midrule
Top-down & random & MLE &&& 31.55 & 97.97 & 73.40 & 33.11 & 54.26 & 18.99 & 25.93 & 101.63\\
Top-down & random & MLE + $\gamma H$ &  & & 30.97 & 96.95 & 72.68 & 32.15 & 54.02 & 18.69 & 25.72 & 99.77\\
\midrule
\midrule
Top-down & random & Tok & Glove sim && 31.47 & 97.72 & 73.49 & 32.93 & 54.38 & 18.75 & 25.88 & 102.33\\
Top-down & random & Tok & Glove sim $r^{\freq}$ && 31.38 & 98.27 & 74.01 & 33.25 & 54.53 & 19.02 & 25.99 & 102.81\\
\midrule
\midrule
Top-down & random & Seq & Hamming & $\V$ & 31.21 & 97.40 & 73.12 & 32.71 & 54.11 & 19.03 & 25.91 & 101.25\\
Top-down & random & Seq & Hamming & $\V_{batch}$ & 31.60 & 98.34 & 73.26 & \bf 32.73 & 54.28 & 18.97 & 25.92 & 101.90\\
Top-down & random & Seq & Hamming & $\V_{refs}$ & 31.58 & 98.46 & \bf 73.53 & 32.59 & 54.36 & 18.91 & 25.98 & \bf 102.33\\
\midrule
Top-down & random & Seq, lazy & Hamming & $\V$ & 31.39 & 98.10 & 73.29 & 32.81 & 54.26 & 18.84 & 25.86 & 101.58\\
Top-down & random & Seq, lazy & Hamming & $\V_{batch}$ & 31.72 & 98.21 & 73.43 & 32.95 & 54.35 & 18.91 & 25.89 & \bf 102.03\\
Top-down & random & Seq, lazy & Hamming & $\V_{refs}$ & 31.82 & 98.35 & \bf 73.53 & \bf 33.09 & 54.31 & 18.78 & 25.74 & 101.89\\
\midrule
Top-down & random & Seq & \cid & $\V$ & 31.16 & 98.16 & 73.08 & 32.51 & 54.12 & 18.91 & 25.89 & 101.84\\
Top-down & random & Seq & \cid & $\V_{batch}$ & 31.27 & 97.83 & \bf 73.50 & \bf 33.04 & 54.48 & 19.15 & 26.13 & \bf 102.98\\
Top-down & random & Seq & \cid & $\V_{refs}$ & 31.67 & 98.57 & 73.42 & 32.91 & 54.52 & 18.98 & 26.00 & 102.23\\
\midrule
Top-down & random & Seq, lazy & \cid & $\V$ & 31.39 & 97.91 & 73.55 & \bf 33.19 & 54.53 & 19.12 & 26.11 & \bf 102.94\\
Top-down & random & Seq, lazy & \cid & $\V_{batch}$ & 31.53 & 97.91 & 73.18 & 32.60 & 54.18 & 18.76 & 25.76 & 101.30\\
Top-down & random & Seq, lazy & \cid & $\V_{refs}$ & 31.84 & 98.61 & \bf 73.92 & 33.10 & 54.57 & 19.02 & 25.87 & 102.64\\
\midrule
\midrule
Top-down & random & Tok-Seq & Hamming & $\V$ & 31.38 & 97.15 & 73.68 & 32.87 & 0 & 0 & 0 & 101.11\\
Top-down & random & Tok-Seq & Hamming & $\V_{batch}$ & 31.37 & 98.12 & 73.86 & 33.32 & 54.34 & 18.89 & 25.86 & 102.90\\
Top-down & random & Tok-Seq & Hamming & $\V_{refs}$ & 31.70 & 98.57 & 73.56 & 33.00 & 54.19 & 18.63 & 25.72 & 101.72\\
Top-down & random & Tok-Seq & \cid & $\V$ & 31.90 & 98.25 & 73.31 & 32.40 & 0 & 0 & 0 & 100.33\\
Top-down & random & Tok-Seq & \cid & $\V_{batch}$ & 31.79 & 98.66 & 73.61 & 32.67 & 54.08 & 18.74 & 25.72 & 101.41\\
Top-down & random & Tok-Seq & \cid & $\V_{refs}$ & 31.68 & 98.51 & \bf 74.28 & \bf 33.34 & 54.66 & 19.07 & 26.00 & \bf 103.81\\
\bottomrule
\end{tabular}
}
\caption{\coco's test set evaluation measures.\label{tab:cap-offline}}
\end{table*}

\section{Experimental evaluation}\label{sec:exp}
In this section, we compare sequence prediction models trained with maximum likelihood (MLE) with our token and sequence-level loss smoothing on two different tasks: image captioning and machine translation.

\subsection{Image captioning}\label{sec:captioning}

\subsubsection{Experimental setup.}\label{para:capsetup}
We use the \coco datatset \citep{lin14eccv}, which consists of 82k training images each annotated with five captions. We use the standard splits of \citet{karpathy15cvpr},  with 5k images for validation, and 5k for test. 
The test set results are generated via  beam search (beam size 3) and are evaluated with the \coco captioning evaluation tool.
We report \cid and \bleu{} scores on this internal test set.
We also report results obtained on the official \coco server that additionally measures \textsc{METEOR} \citep{meteor} and \textsc{ROUGE-L} \citep{lin04acl}.
\begin{table*}
\centering
\resizebox{\textwidth}{!}{
\begin{tabular}{lcccccccccccccccc}
\toprule
& \multicolumn{2}{c}{\bleu{-1}}   
& \multicolumn{2}{c}{\bleu{-2}}   
& \multicolumn{2}{c}{\bleu{-3}}   
& \multicolumn{2}{c}{\bleu{-4}}   
    & \multicolumn{2}{c}{METEOR}   
    & \multicolumn{2}{c}{ROUGE-L}  
    & \multicolumn{2}{c}{\cid}    
    & \multicolumn{2}{c}{SPICE} \\
\cmidrule(r){2-3} \cmidrule(r){4-5}
\cmidrule(r){6-7}\cmidrule(r){8-9}
\cmidrule(r){10-11}\cmidrule(r){11-13}
\cmidrule(r){14-15}\cmidrule(r){16-17}
    & c5           & c40           
    & c5           & c40           
    & c5           & c40           
    & c5           & c40           
    & c5           & c40           
    & c5           & c40           
    & c5           & c40           
    & c5           & c40         \\
\midrule
    Google NIC\ens~\cite{vinyals15cvpr}&71.3&89.5& 54.2&80.2& 40.7&69.4& 30.9&58.7 &25.4&34.6 &53.0&68.2 &94.3&94.6 &18.2&63.6\\
    Hard-Attention~\cite{xu15icml}&70.5&88.1 &52.8&77.9 &38.3&65.8 &27.7&53.7 &24.1&32.2 &51.6&65.4 &86.5&89.3 &17.2&59.8\\
    ATT-FCN\ens~\cite{you16cvpr}&73.1&90.0 &56.5&81.5 &42.4&70.9 &31.6&59.9 &25.0&33.5 &53.5&68.2 &94.3&95.8 &18.2&63.1\\
    Review Net\ens~\cite{yang16nips}&72.0&90.0 &55.0&81.2 &41.4&70.5 &31.3&59.7 &25.6&34.7 &53.3&68.6 &96.5&96.9 &18.5&64.9\\
    Adaptive\ens~\cite{lu17cvpr}&74.8&92.0 &58.4&84.5 &44.4&74.4 &33.6&63.7 &26.4&35.9 &55.0&70.5 &104.2&105.9 &19.7&67.3\\
    \midrule
    SCST:Att2all\ens\rl~\cite{rennie17cvpr}&78.1&93.7 &61.9&86.0 &47.0&75.9 &35.2&64.5 &27.0&35.5 &56.3&70.7 &114.7&116.7 &-&-\\
    LSTM-A3\ens\rl\add~\cite{yao17iclr}&78.7&93.7&62.7&86.7&47.6&76.5&35.6&65.2&27.0&35.4&56.4&70.5&116&118&-&-\\
    Up-Down\ens\rl\add~\cite{anderson17arxiv} & 80.2 & 95.2 &  64.1 & 88.8  & 49.1 & 79.4  & 36.9 & 68.5 &   27.6 & 36.7 &  57.1 & 72.4 & 117.9 & 120.5  & - & - \\
\midrule
Ours: Tok-Seq \cid    &72.6&89.7&55.7&80.9&41.2&69.8&30.2&58.3&25.5&34.0&53.5&68.0&96.4&99.4&-&-\\ 
Ours: Tok-Seq \cid \ens &74.9&92.4&58.5&84.9&44.8&75.1&34.3&64.7&26.5&36.1&55.2&71.1&103.9&104.2& - & -\\
\bottomrule
\end{tabular}
}
\caption{\coco's server evaluation \label{tab:cap-online}. (\ens) for ensemble submissions, (\rl) for submissions with CIDEr optimization and (\add) for models using additional data.}
\end{table*}
We experiment with both non-attentive LSTMs \citep{vinyals15cvpr} and the ResNet baseline of the state-of-the-art top-down attention \citep{anderson17arxiv}.

The \coco vocabulary consists of 9,800 words that occur at least 5 times in the training set. Additional details and hyperparameters can be found in \app{capsetup}.
\subsubsection{Results and discussion}
\paragraph{Restricted vocabulary sampling}\label{para:capsubset}

In this section, we evaluate the impact of the vocabulary subset from which we sample the modified sentences for sequence-level smoothing. 
We experiment with two rewards: {\cid}, which scores \wrt all five available reference sentences, and Hamming distance reward taking only a single reference into account. For each reward we train our (Seq) models with each of the three subsets detailed previously in \parref{subset}.

From the results in \tab{cap-offline} we note that for the inattentive models, sampling from $\V_{refs}$ or $\V_{batch}$ has a better performance than sampling from the full vocabulary on all metrics. In fact, using these subsets introduces a useful bias to the model and improves performance. 
This improvement is most notable using the  \cid reward that scores candidate sequences \wrt to multiple references, which stabilizes the scoring of the candidates.

With an attentive decoder, no matter the reward, re-sampling sentences with words from $\V_{ref}$ rather than the full vocabulary $\V$ is better for both reward functions, and all metrics. 
Additional experimental results, presented in \app{subset-results},
obtained with a \bleuf reward, in its single and multiple references variants, further corroborate this conclusion.

\paragraph{Lazy training.}
From the results of \tab{cap-offline}, we see that lazy sequence-level smoothing is competitive with exact non-lazy sequence-level smoothing, while requiring roughly equivalent training time as MLE. We provide detailed timing results in \app{speed}.

\paragraph{Overall}
For reference, we include in \tab{cap-offline} baseline results obtained using MLE, and our implementation of MLE with entropy regularization (MLE+$\gamma H$) \citep{pereyra17iclr}, as well as the RAML approach of \citet{norouzi16nips} which corresponds to sequence-level smoothing based on the Hamming reward and sampling replacements from the full vocabulary (Seq, Hamming, $\mathcal V$)

We observe that entropy smoothing is not able to improve performance much over MLE for the model without attention, and even deteriorates for the attention model. 
We improve upon RAML by choosing an adequate subset of vocabulary for substitutions.

We also report the performances of token-level smoothing, where the promotion of rare tokens boosted the scores in both attentive and non-attentive models. 

For sequence-level smoothing, choosing a task-relevant reward with importance sampling yielded better results than plain Hamming distance.

Moreover, we used the two smoothing schemes (Tok-Seq) and achieved the best results with $\cid$ as a reward for sequence-level smoothing combined with a token-level smoothing that promotes rare tokens improving \cid from 93.59 (MLE) to 99.92 for the model without attention, and improving from 101.63 to 103.81 with attention.  

\paragraph{Qualitative results.}
\begin{figure*}[!ht]
\begin{tabular}{ll}
\includegraphics[width=.49\textwidth]{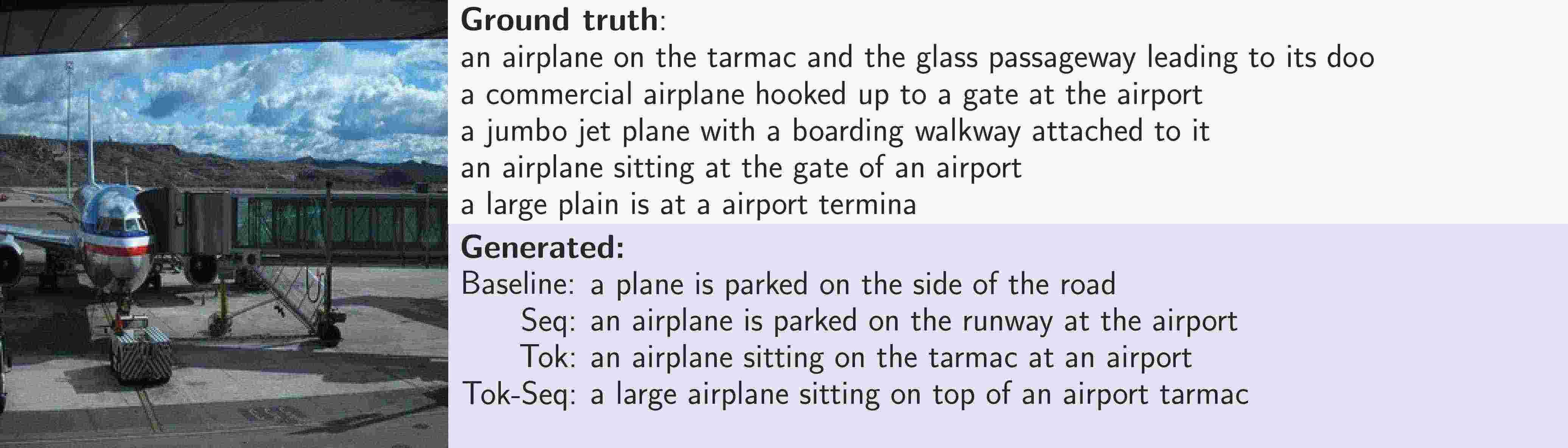} &
\includegraphics[width=.49\textwidth]{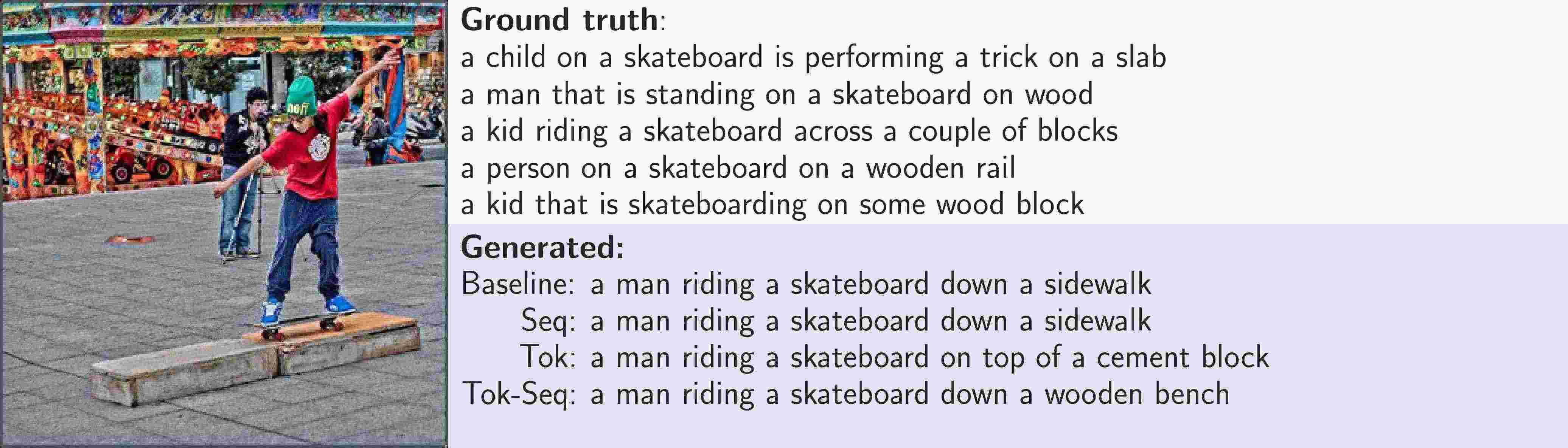}\\
\includegraphics[width=.49\textwidth]{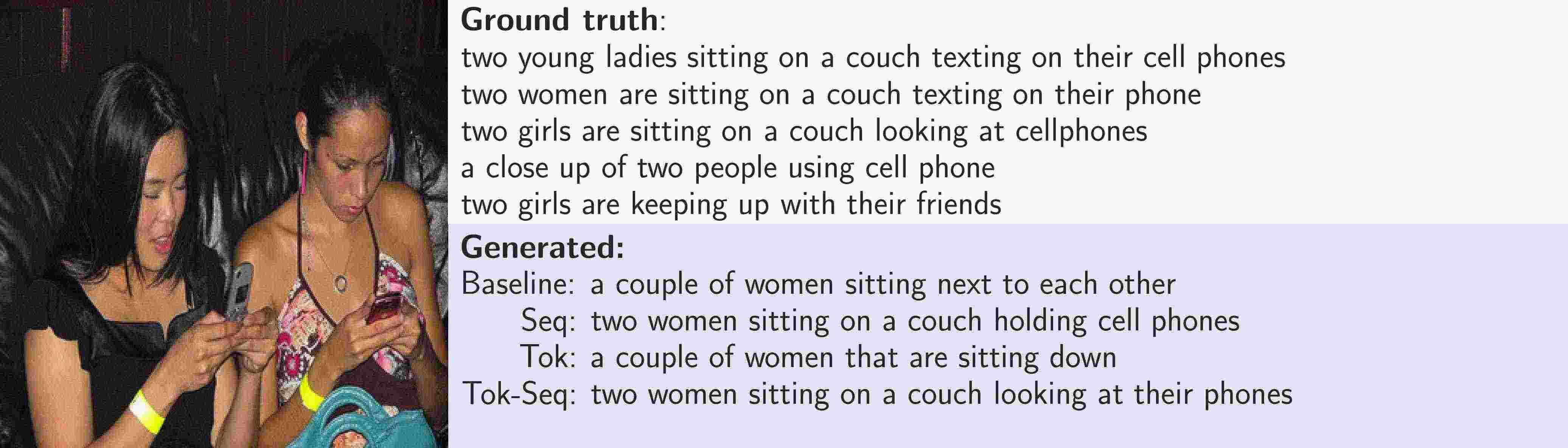} &
\includegraphics[width=.49\textwidth]{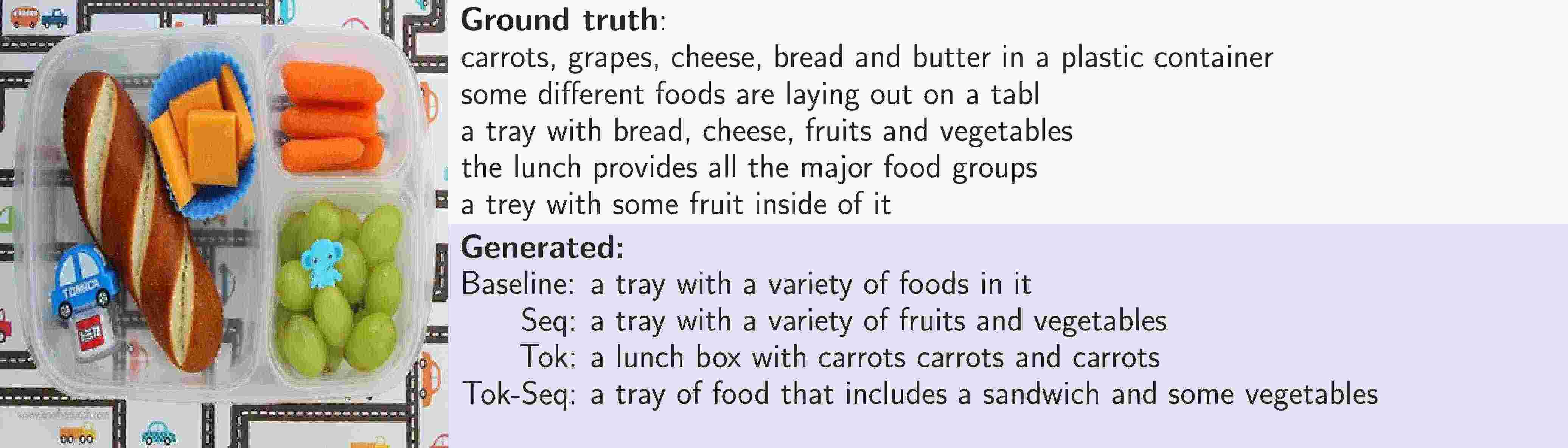} \\
\end{tabular}
\caption{Examples of generated captions with the baseline MLE and our models with attention.\label{fig:exhibits}}
\end{figure*}
In \fig{exhibits} we showcase captions obtained with MLE and our three variants of smoothing \ie token-level (Tok), sequence-level (Seq) and the combination (Tok-Seq). We note  that the sequence-level smoothing tend to generate lengthy captions overall, which is maintained in the combination. On the other hand, the token-level smoothing allows for a better recognition of objects in the image that stems from the robust training of the classifier \eg the 'cement block' in the top right image or the carrots in the bottom right.
More examples are available in \app{caps_examples}

\paragraph{Comparison to the state of the art.}
We compare our model to state-of-the-art systems on the \coco evaluation server in \tab{cap-online}.
We submitted a single model (Tok-Seq, \cid, $\V_{refs}$) as well as an ensemble of five models with different initializations trained on the  training set plus 35k images from  the dev set (a total of 117k images) to the \coco server. 
The three best results on the server \citep{rennie17cvpr, yao17iclr, anderson17arxiv} are trained in two stages where they first train using MLE,  before switching to  policy gradient methods based on CIDEr. 
\citet{anderson17arxiv} reported an increase of 5.8\% of \cid on the test split after the \cid optimization.
Moreover, \citet{yao17iclr} uses additional information about image regions to train the attributes classifiers, while \citet{anderson17arxiv} pre-trains its bottom-up attention model on the Visual Genome dataset \citep{krishna17ijcv}.
\citet{lu17cvpr, yao17iclr} use the same CNN encoder as ours (ResNet-152), \citep{vinyals15cvpr, yang16nips} use Inception-v3 \citep{szegedy16cvpr} for image encoding and \citet{rennie17cvpr, anderson17arxiv} use Resnet-101, both of which have similar performances to ResNet-152 on ImageNet classification \citep{cnn-benchmark}. 
\subsection{Machine translation}
\subsubsection{Experimental setup.}
For this task we validate the effectiveness of our approaches on two different datasets. The first is WMT'14 English to French,
in its filtered version, with 12M sentence pairs obtained after dynamically selecting a “clean” subset of 348M words out of the original “noisy” 850M words \citep{bahdanau15iclr, cho14emnlp, sutskever14nips}.
The second benchmark is IWSLT'14 German to English consisting of around 150k pairs for training. 
In all our experiments we use the attentive model of \citep{bahdanau15iclr}
The hyperparameters of each of these models as well as any additional pre-processing can be found in \app{nmtsetup} 

To assess the translation quality we report the \bleuf metric.

\subsubsection{Results and analysis}
\begin{table}[!ht]
\centering
\resizebox{\columnwidth}{!}{
\begin{tabular}{p{1.3cm}ccHcc}
    Loss & Reward & $\V_{sub}$ & Beam  & \bf WMT'14 & \bf IWSLT'14 \\
\midrule
MLE &  &  & 5 & 30.03 & 27.55 \\
\midrule
tok & Glove sim &  & 5 & 30.16 & 27.69\\
tok & Glove sim $r^{freq}$ &  & 5 & 30.19  & 27.83\\
\midrule
Seq& Hamming & $\V$ & 5 & 30.85 &  27.98\\
Seq & Hamming & $\V_{batch}$ & 5 & 31.18  & 28.54\\ 
Seq & \bleu{-4} & $\V_{batch}$ & 5 & 31.29  & 28.56\\
\midrule
Tok-Seq & Hamming & $\V_{batch}$ & 5 & 31.36   & 28.70\\
Tok-Seq & \bleu{-4} & $\V_{batch}$ & 5 & 31.39  & 28.74\\
\bottomrule
\end{tabular}
}
\caption{Tokenized BLEU score on WMT'14 En-Fr evaluated on the news-test-2014 set. And Tokenzied, case-insensitive BLEU on IWSLT'14 De-En.}

\label{tab:nmt}
\end{table}
We present our results in \tab{nmt}. On both benchmarks, we improve on both MLE and RAML approach of \citet{norouzi16nips}
 (Seq, Hamming, $\V$): using the smaller batch-vocabulary for replacement improves results, and using importance sampling based on \bleuf further boosts results.
In this case, unlike in the captioning experiment, token-level smoothing brings smaller improvements.
The combination of both smoothing approaches gives best results, similar to what was observed for image captioning, improving the MLE \bleuf from 30.03 to 31.39 on WMT'14 and from 27.55 to 28.74 on IWSLT'14.
The outputs of our best model are compared to the MLE in some examples showcased in \app{nmt}.
\section{Conclusion}
We investigated the use of loss smoothing approaches to improve over maximum likelihood estimation of RNN language models. 
We generalized the  sequence-level smoothing RAML approach of \citet{norouzi16nips} to the token-level by smoothing the ground-truth target across semantically similar tokens. For the sequence-level, which is computationally expensive, we introduced an efficient ``lazy'' evaluation scheme, and introduced an improved re-sampling  strategy. 
Experimental evaluation on image captioning and machine translation demonstrates the complementarity of sequence-level and token-level loss smoothing, improving  over both the maximum likelihood and RAML.

\medskip
{{\bf Acknowledgment.}
This work has been partially supported by the grant ANR-16-CE23-0006 ``Deep in
France'' and LabEx PERSYVAL-Lab (ANR-11-LABX-0025-01). 
}
\bibliography{refs,jjv}
\bibliographystyle{acl_natbib}
\appendix
\newpage
\onecolumn
\newpage
\onecolumn
\section{Stratified sampling of the Hamming distance}\label{appendix:strat}
\newcite{norouzi16nips} detail the steps to drawing samples from the reward  distribution based on the edit distance with stratified sampling.
In this section we show how we sample from the Hamming distance (a special case of the edit distance) reward and how we handle large vocabularies to generate reasonable candidates.
To draw from the Hamming distance reward, we proceed as follows:
\begin{enumerate}
    \compresslist
    \item  Sample a distance $d$ from $\{0,\ldots, T\}$.
    \item  Pick $d$ positions in the sequence to be changed among $\{1,\ldots, T\}$.
    \item  Sample substitutions from a subset $\V_{sub}$ of the vocabulary ($|\V_{sub}| = V_{sub}$).
\end{enumerate}
To sample a distance, we partition the set of sequences in $\V_{sub}$ terms $\V_{sub}^T$ with respect to their distance to the ground truth $\gt{}$:
\[\begin{cases}
    S_d = \{y\in\V_{sub}^T |\, \ham(y, \gt{})=d\},\\
    \V_{sub}^T = \underset{d}\cup S_d,\\
    \forall d,d': S_d\cap S_{d'} = \emptyset.
\end{cases}\]

To each distance $d$ in $\{0,...,T\}$ we assign the portion of rewards covered by $S_d$ in $\V_{sub}^T$.  Since all elements of $S_d$ are assigned the same reward $e^{-\frac{d}{\tau}}$; we need only to multiply it by the set's size $|S_d|$. Counting the elements of $|S_d|$ is straight-forward: we choose $d$ elements to alter in $\gt{}$ $\left({T \choose d} \text{combinations}\right)$ and at each position we have $(V_{sub}-1)$ possibilities.
The sampling distribution is obtained as:

\begin{align}
    p(d)& = \rt(S_d) / \rt(\V_{sub}^T) \\
        & =\frac{\sum\limits_{y \in S_d} \rt(y|\gt{})}{\sum\limits_{y \in \V_{sub}^T} \rt(y|\gt{})}.
\end{align}
Given that $\{S_d\}_d$ form a partition of $\V_{sub}^T$,
\begin{align}
    \sum_{y\in\V_{sub}^T}r(y|\gt{}) & = \sum_d\sum_{y\in S_d} e^{-\frac{d}{\tau}}\\
                                    & = \sum_d {T\choose d}(V_{sub}-1)^d e^{-\frac{d}{\tau}}\\
                                    & = \left( (V_{sub}-1)e^{-\frac{1}{\tau}}+1\right)^T,
\end{align}
we find:
\begin{equation}
    p(d) = {T \choose d} \frac{(V_{sub}-1)^d e^{-\frac{d}{\tau}}}{((V_{sub}-1)e^{-\frac{1}{\tau}} + 1)^T}.
\end{equation}

\section{Captioning}
\subsection{Experimental setup}\label{appendix:capsetup}
Out of  vocabulary words are replaced by \textsc{<unk>} token and the captions longer than 16 words are truncated.
As image encoding, we average-pool the features in the  last convolutional layer of ResNet-152 \citep{he16cvpr} pre-trained on ImageNet. 
The 2048-dimensional image signature is further mapped to $\R^{512}$ to fit the word-embedding dimension, so it can be used as the first token fed to the RNN decoder.
We use a single-layer RNN with $d=512$ LSTM  units.

For optimization, we use Adam \citep{kingma15iclr} with a batch size of 10 images, \ie 50 sentences. We follow \citet{lu17cvpr,perdersoli17iccv} and train in two stages: the first, optimizing the language model alone with an initial learning rate of 5e-4 annealed by a factor of 0.6 every 3 epochs starting from the 5th one. We train for up to 20 epochs with early stopping if \cid score on  the validation set  does not improve.
In the second stage, we optimize the language model conjointly with $conv_4$ and $conv_5$ (the last 39 building blocks of ResNet-152) of the CNN model. The initial learning rate is of 6e-5 and diminishes by a factor of 0.8 every 4 epochs. The same early-stopping strategy is applied.
For the token-level reward, we use GloVe \citep{pennington14emnlp} as our word embedding, 
which we train on the captions in the 
 \coco training set. 
In preliminary experiments  using the publicly available 300-dimensional GloVe vectors trained on Wikipedia 2014 + Gigaword 
worsens the model's results.

\subsection{Restricted vocabulary sampling - supplementary results}\label{appendix:subset-results}

In \tab{seqsmooth} we provide additional results for sequence-level smoothing, 
when using the \bleuf as reward function. We include results when computing \bleuf \wrt all reference sequences (like also done for \cid), and when computing \wrt a single randomly selected reference sentence (as is the case for Hamming). 
With the \bleuf reward, using $\V_{refs}$ yields best results in all but a single case.
This underlines the effectiveness of sampling replacement words that are relevant to the task in sequence-level smoothing.

\begin{table}[H]
\begin{minipage}[t]{.5\textwidth}
\centering
\resizebox{\textwidth}{!}{
\begin{tabular}{HllccHc}
\toprule
\multicolumn{7}{c}{\bf Captioning without attention}\\
\midrule
Loss & Reward & $\V_{sub}$ & BLEU-1 & \bleu{-4} & METEOR & $\cid$\\
\midrule
ML &  Dirac &  & 70.63 & 30.14 & 24.75 & 93.59\\
\midrule
Seq & Hamming & $\V$ & 71.76 & 31.16 & 25.19 & 96.37\\
Seq & Hamming & $\V_{batch}$ & 71.46 & 31.15 & 25.13 & \bf 96.53 \\
Seq & Hamming & $\V_{refs}$ & \bf 71.80 & \bf 31.63 & 25.05 & 96.22  \\
\midrule
Seq & \bleu{-4} single & $\V$ & 71.30 & 31.11 & 25.04 & 96.11 \\
Seq & \bleu{-4} single & $\V_{batch}$ & 71.13 & 30.84 & 24.82 & 94.74 \\
Seq & \bleu{-4} single & $\V_{refs}$ & \bf 71.78 & \bf 31.63 & 25.21 & \bf 97.29 \\
\midrule
Seq & \bleu{-4}  & $\V$ & 71.56 & \bf 31.47 & 25.26 & 96.56 \\
Seq & \bleu{-4}  & $\V_{batch}$ & 71.41 & 30.87 & 25.11 & 95.69 \\
Seq & \bleu{-4}  & $\V_{refs}$ & \bf 72.08 & 31.41 & 25.23 & \bf 97.21 \\
\midrule
Seq & $\cid$ & $\V$ & 71.05 & 30.46 & 24.86 & 94.40  \\
Seq & $\cid$ & $\V_{batch}$ & 71.51 & 31.17 & 25.02 & 95.78 \\
Seq & $\cid$ & $\V_{refs}$ & \bf 71.93 & \bf 31.41 & 25.08 & \bf 96.81  \\
\bottomrule
\end{tabular}
}
\end{minipage}\hfill
\begin{minipage}[t]{.5\textwidth}
\resizebox{\textwidth}{!}{
\begin{tabular}{HllccHc}
\toprule
\multicolumn{7}{c}{\bf Captioning with attention}\\
\midrule
Loss & Reward & $\V_{sub}$ & BLEU-1 & \bleu{-4} & METEOR & $\cid$\\
\midrule
ML & Dirac & & 73.40 & 33.11 & 25.93 & 101.63\\
\midrule
Seq & Hamming & $\V$ & 73.12 & 32.71 & 25.91 & 101.25 \\
Seq & Hamming & $\V_{batch}$ & 73.26 & \bf 32.73 & 25.92 & 101.90 \\
Seq & Hamming & $\V_{refs}$ & \bf 73.53 & 32.59 & 25.98 & \bf 102.33  \\
\midrule
Seq & \bleu{-4} single & $\V$ & 72.98 & 32.55 & 25.85 & 101.15 \\
Seq & \bleu{-4} single & $\V_{batch}$ & 72.98 & 32.48 & 25.89 & 101.05 \\
Seq & \bleu{-4} single & $\V_{refs}$ & \bf 73.41 & \bf 32.69 & 25.83 & \bf 101.35 \\
\midrule
Seq & \bleu{-4}  & $\V$ 		& 73.39 & 32.89 & 26.10 & 102.60\\
Seq & \bleu{-4}  & $\V_{batch}$ & 73.24 & 32.74 & 26.07 & 102.49\\
Seq & \bleu{-4}  & $\V_{refs}$  & \bf 73.55 & \bf 33.03 & 26.06 & \bf 102.72 \\
\midrule
Seq & $\cid$ & $\V$ & 73.08 & 32.51 & 25.89 & 101.84  \\
Seq & $\cid$ & $\V_{batch}$ & \bf 73.50 & \bf 33.04 & 26.13 & \bf 102.98  \\
Seq & $\cid$ & $\V_{refs}$ & 73.42 & 32.91 & 26.00 & 102.23  \\
\bottomrule
\end{tabular}
}
\end{minipage}
\caption{Captioning performance on MSCOCO when training with sequence-level loss smoothing.}
\label{tab:seqsmooth}
\end{table}

\subsection{Training time}\label{appendix:speed}
We report below (\tab{time}) the average wall time to process a single batch (10 images \ie 50 captions) when training the RNN language model with fixed CNN (without attention) on a Titan X GPU. We can clearly see that the lazy training is faster compared to the standard sequence smoothing and that the token-level smoothing does not hinder the training speed. 
\begin{table*}[!ht]
    \centering
    \resizebox{\textwidth}{!}{
    \begin{tabular}{|c|c|c|*{9}{c}|}
      \toprule
      Loss & MLE & Tok & Seq & Seq lazy & Seq & Seq lazy & Seq & Seq lazy & Tok-Seq & Tok-Seq & Tok-Seq\\
      \midrule
      Reward & & Glove sim & \multicolumn{9}{c|}{Hamming} \\
      \midrule
      $\V_{sub}$ & & & $\V$ & $\V$ & $\V_{batch}$ & $\V_{batch}$  & $\V_{refs}$  & $\V_{refs}$ &  $\V$ & $\V_{batch}$ &  $\V_{refs}$\\ 
      ms/batch  & 347 & 359 & 390    & 349 & 395 & 337   & 401 & 336 & 445  & 446 & 453     \\ 
       \bottomrule
    \end{tabular}
    }
    \caption{Average training time per batch for different losses}\label{tab:time}
\end{table*}

\newpage
\subsection{Additional examples}\label{appendix:caps_examples}
\begin{longtable}{ll}
\includegraphics[width=.49\textwidth]{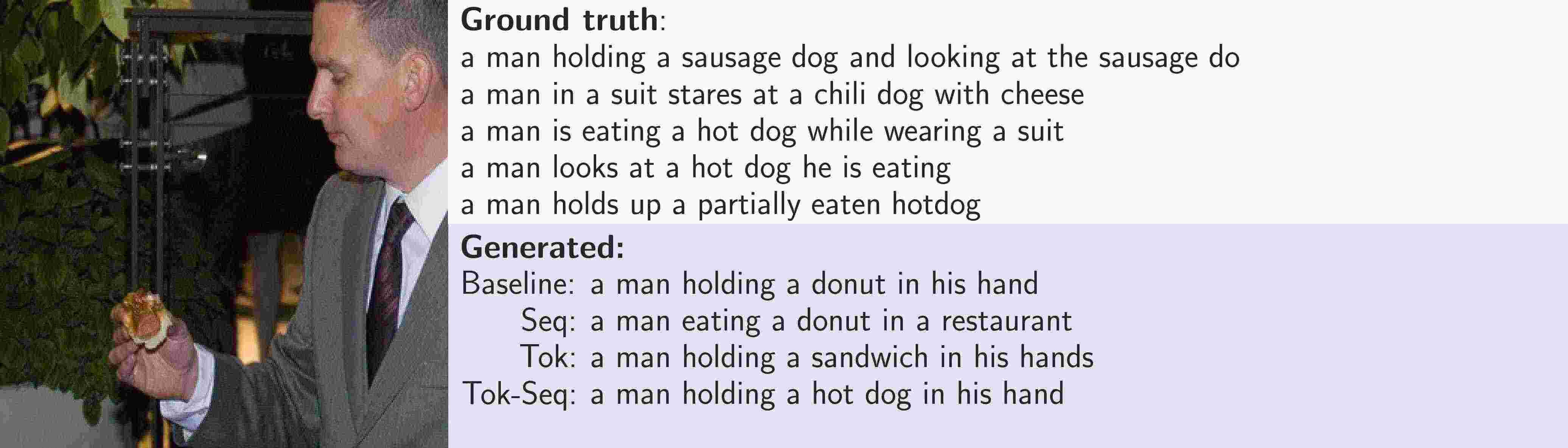} & \includegraphics[width=.49\textwidth]{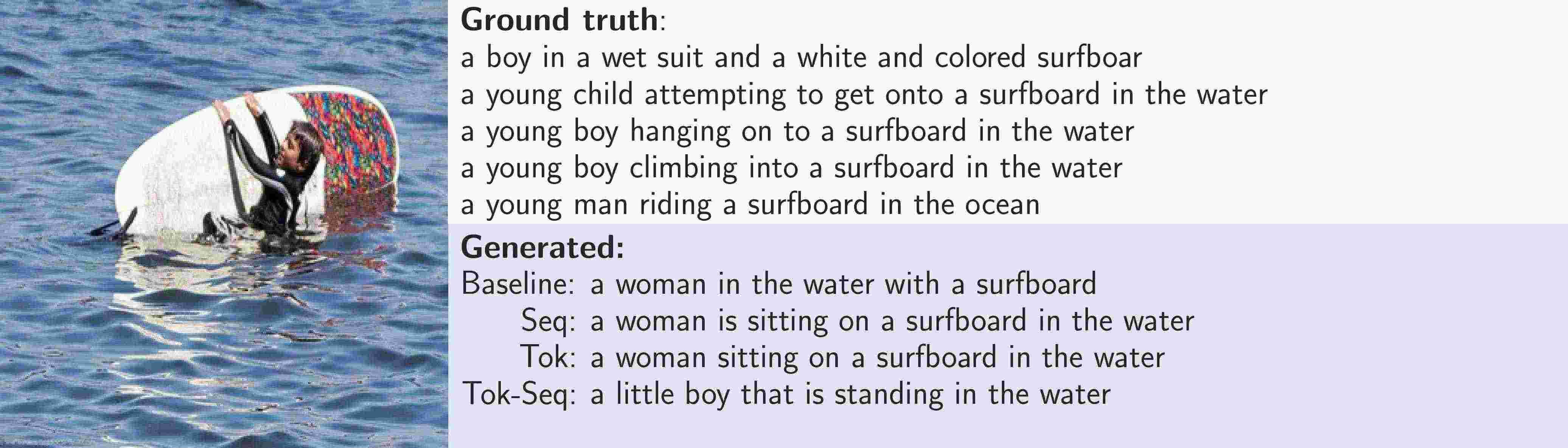} \\
\includegraphics[width=.49\textwidth]{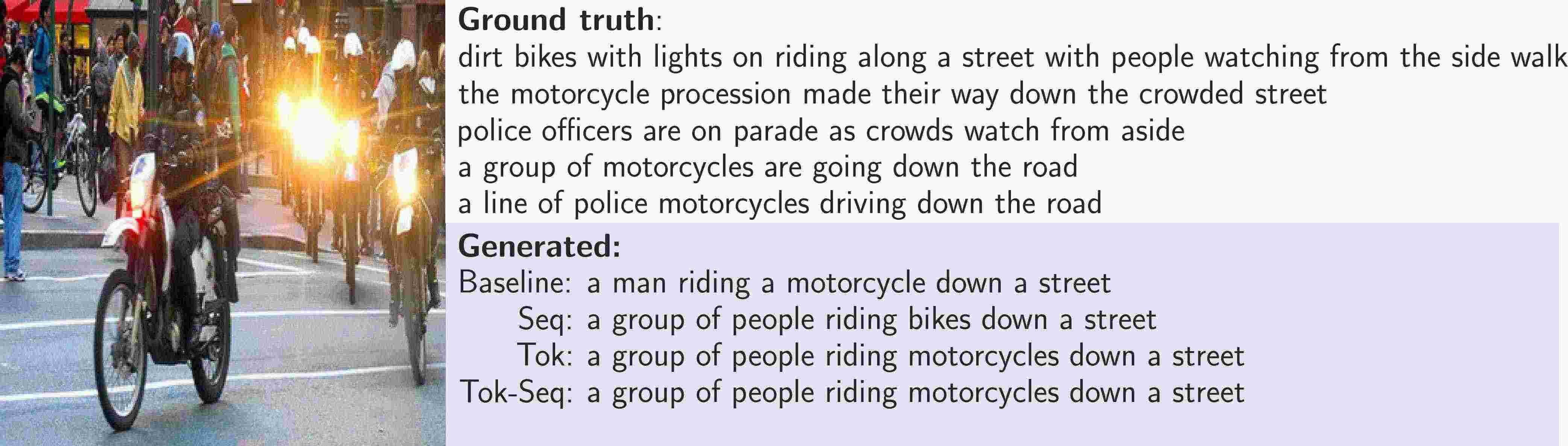} & \includegraphics[width=.49\textwidth]{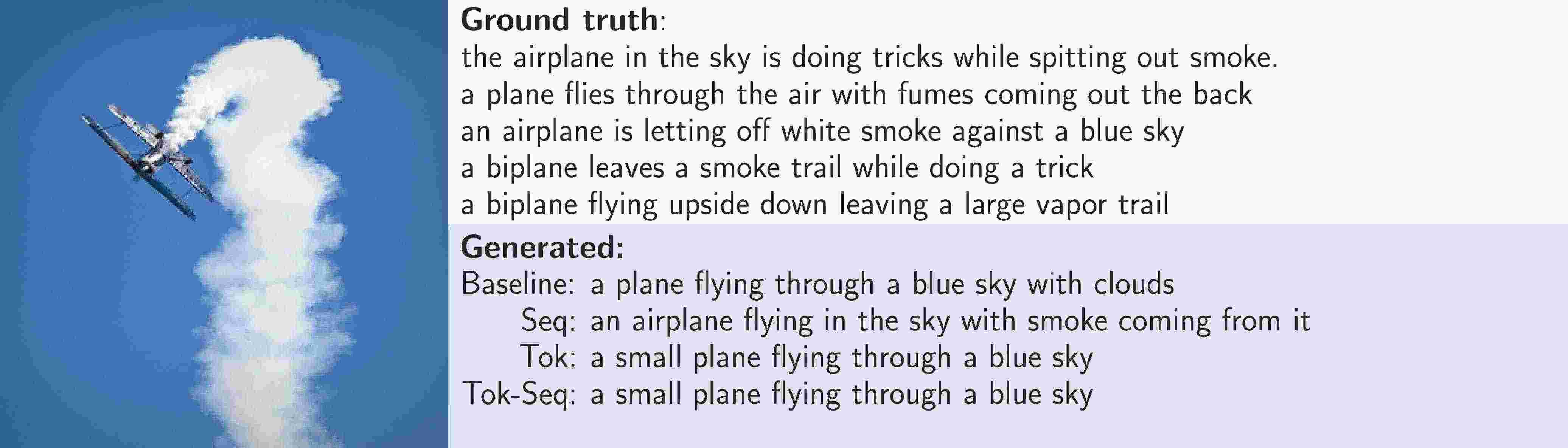} \\
\includegraphics[width=.49\textwidth]{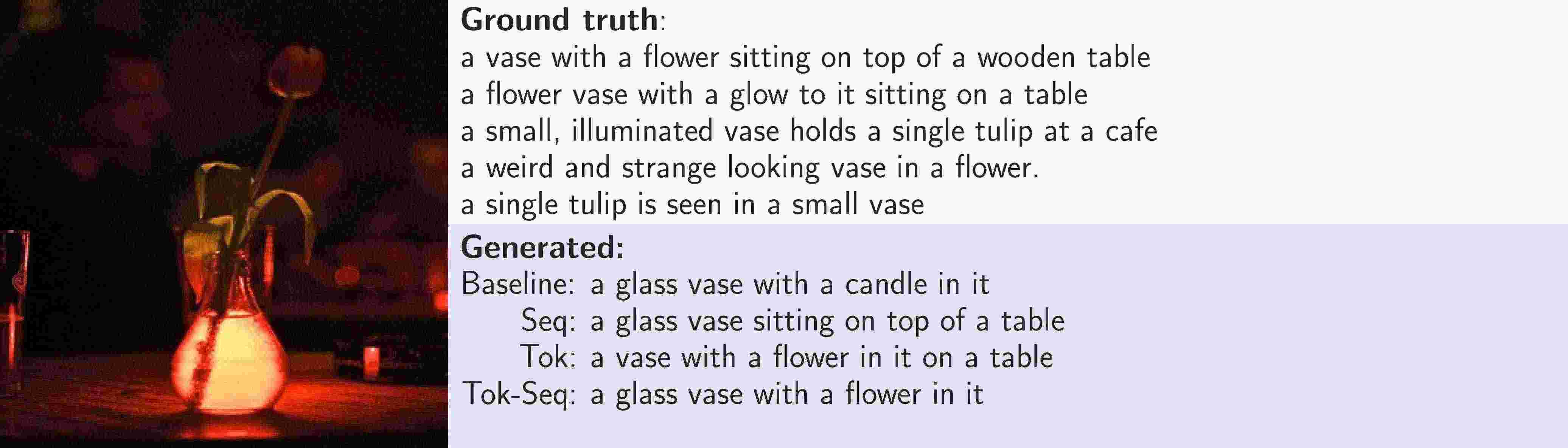} & \includegraphics[width=.49\textwidth]{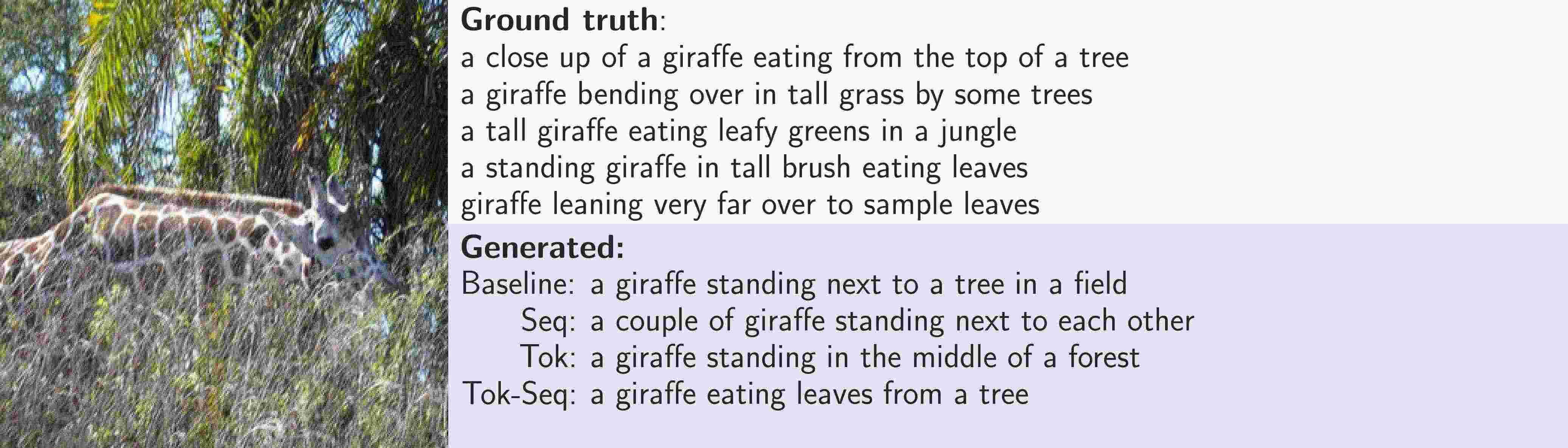} \\
\includegraphics[width=.49\textwidth]{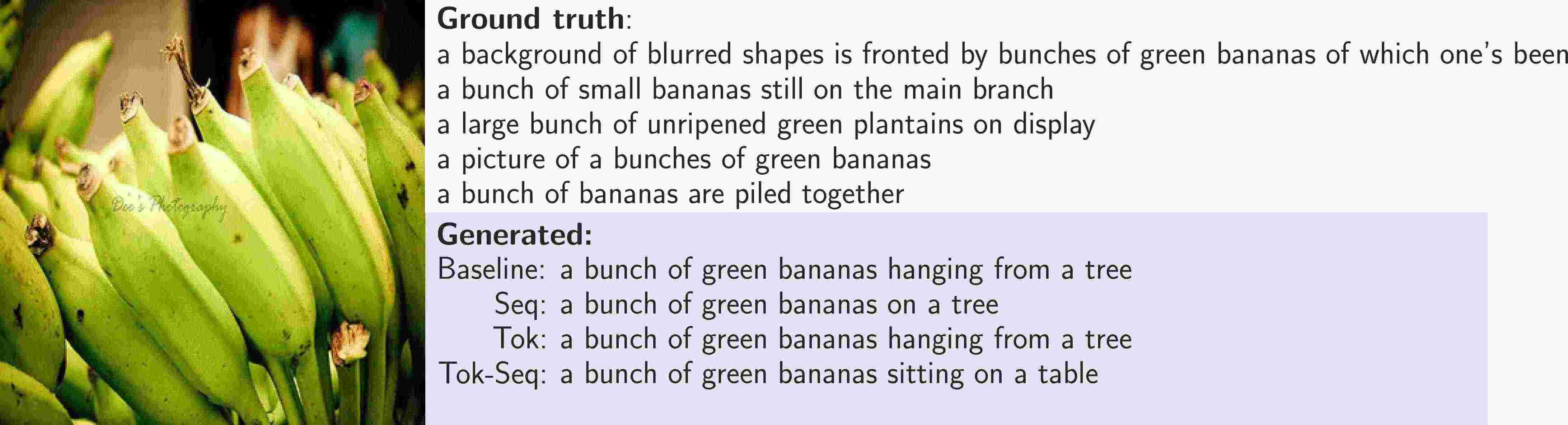} & \includegraphics[width=.49\textwidth]{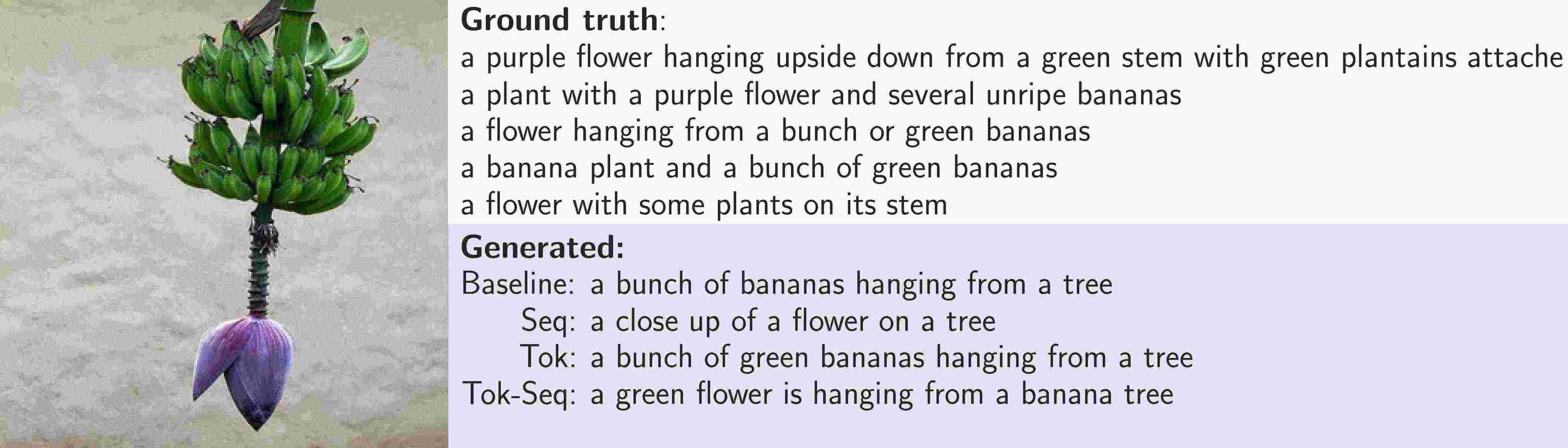} \\
\includegraphics[width=.49\textwidth]{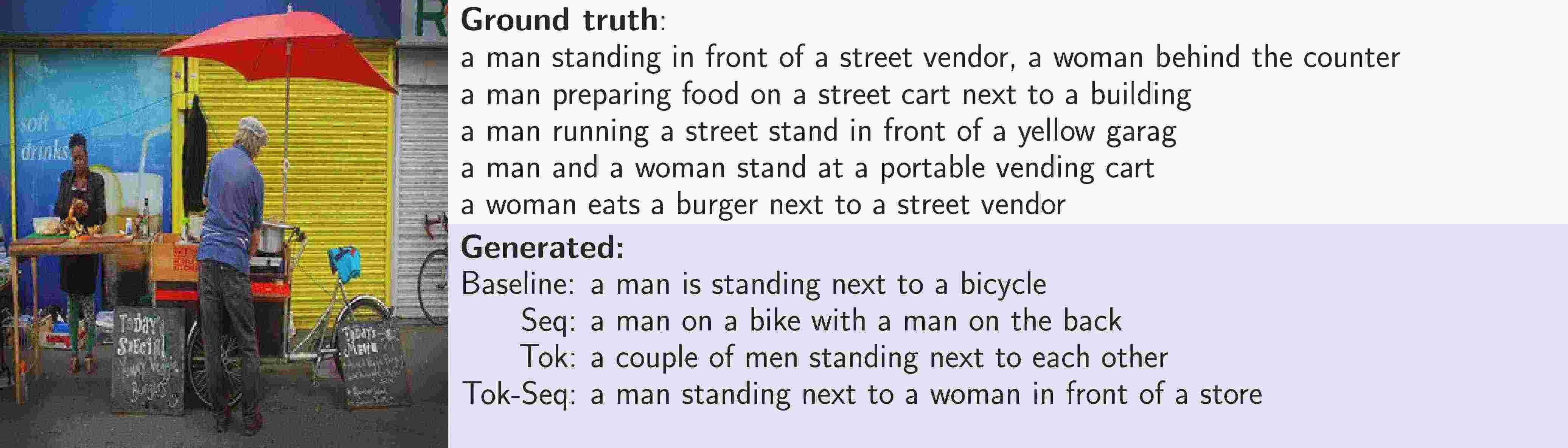} & \includegraphics[width=.49\textwidth]{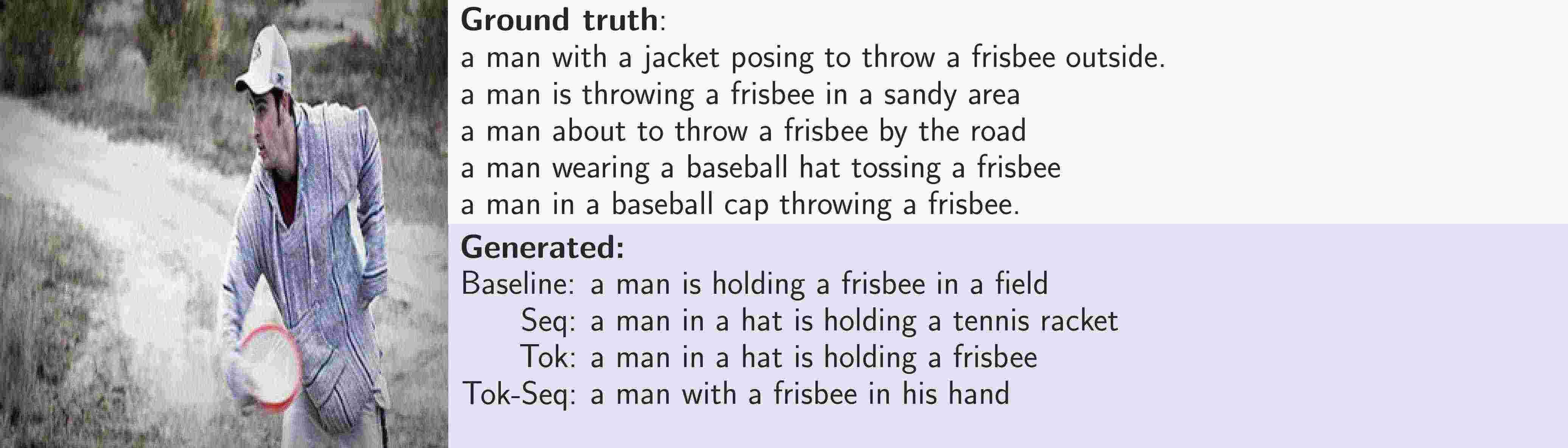} \\
\includegraphics[width=.49\textwidth]{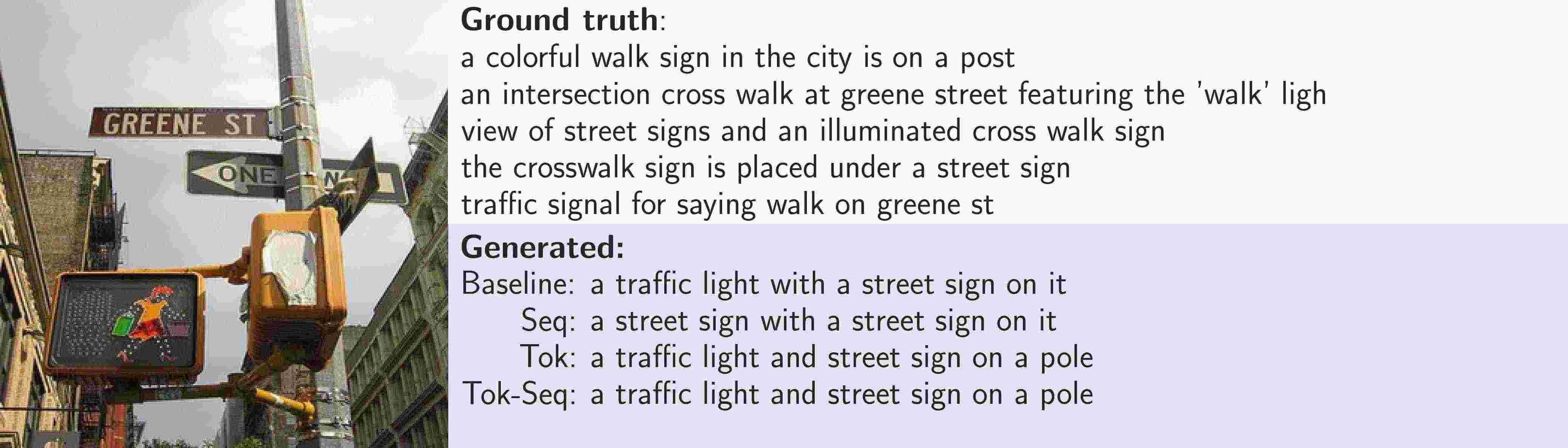} & \includegraphics[width=.49\textwidth]{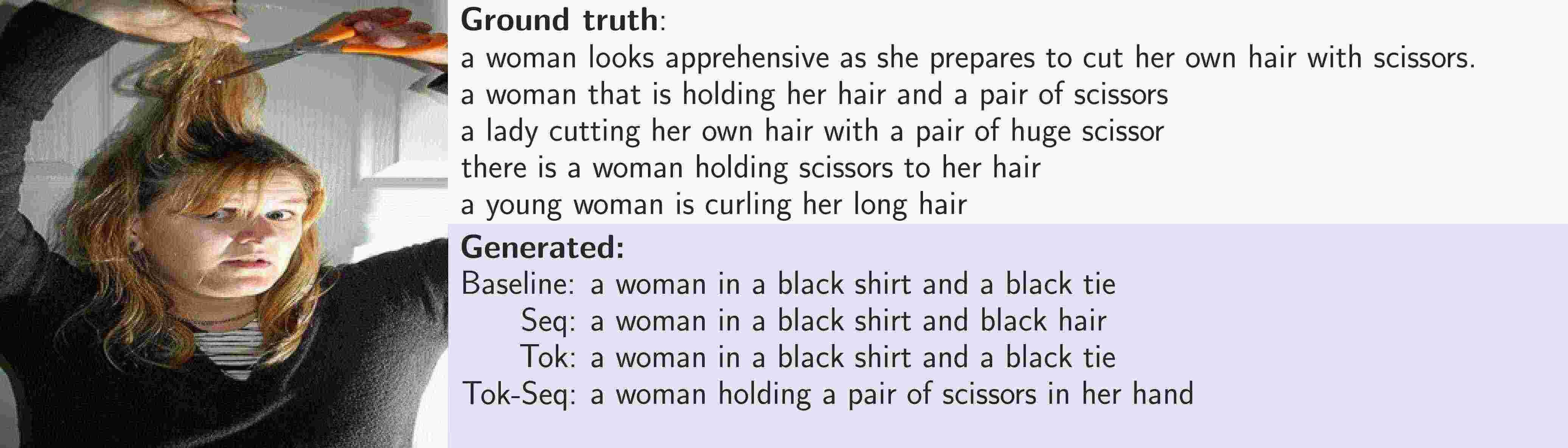} \\
\includegraphics[width=.49\textwidth]{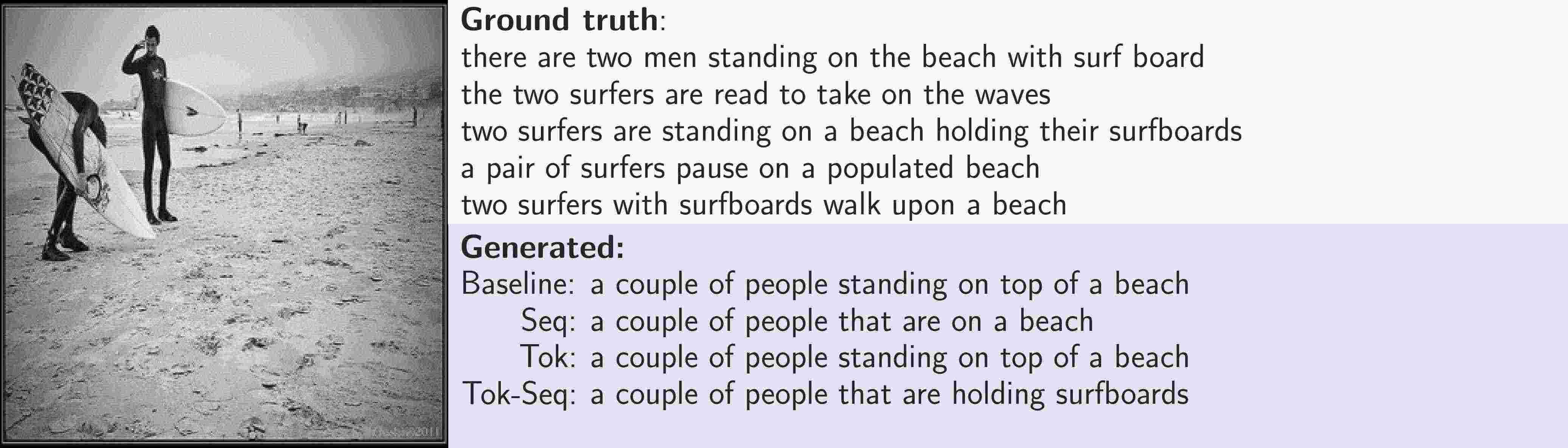} & \includegraphics[width=.49\textwidth]{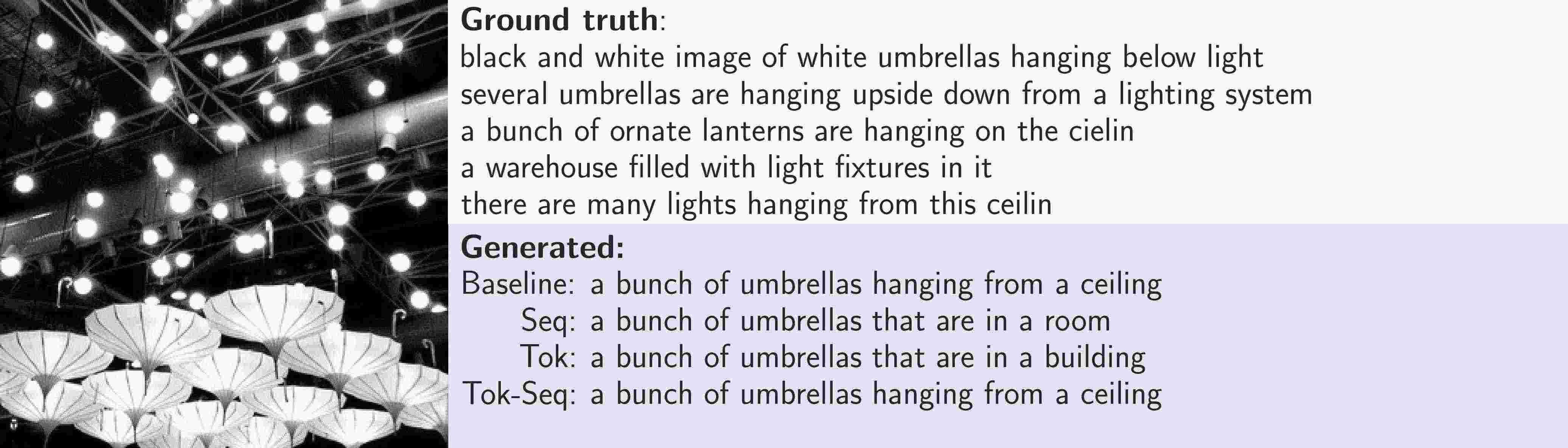} \\
\includegraphics[width=.49\textwidth]{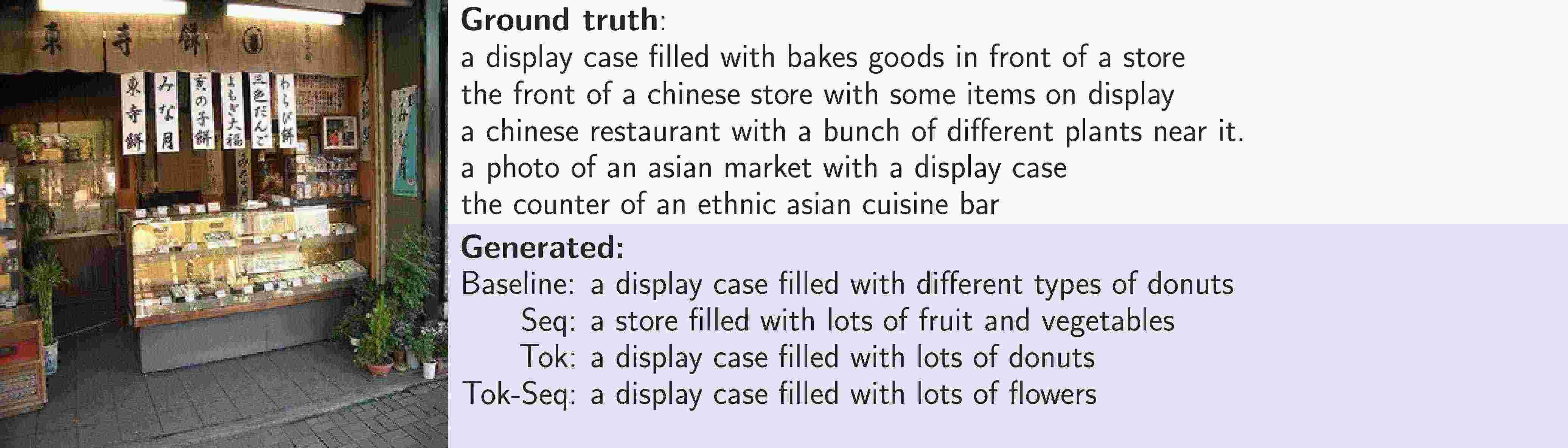} & \includegraphics[width=.49\textwidth]{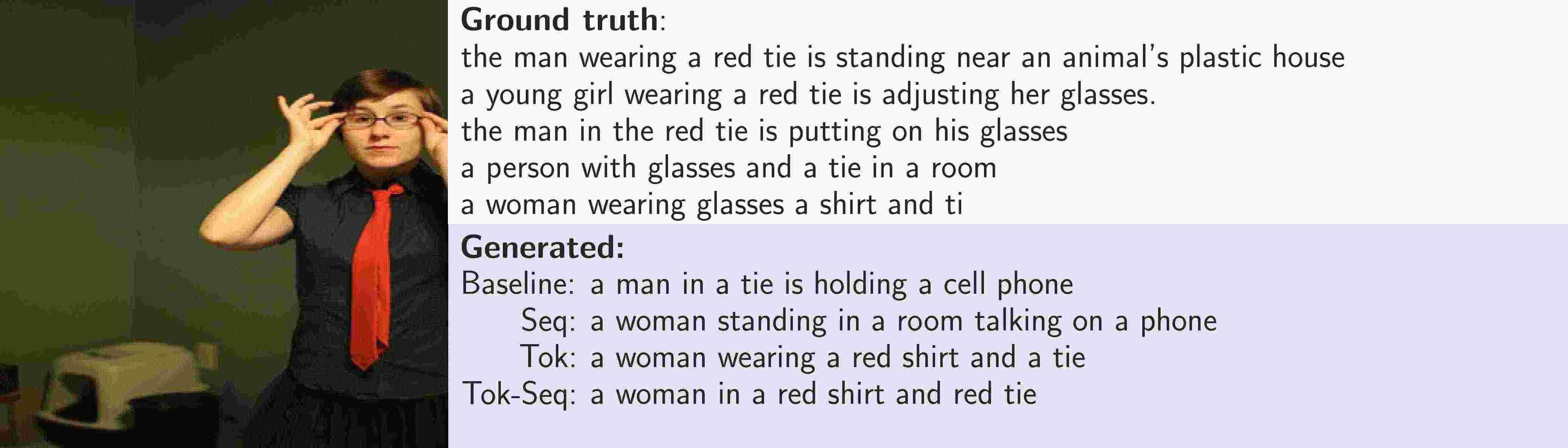} \\
\includegraphics[width=.49\textwidth]{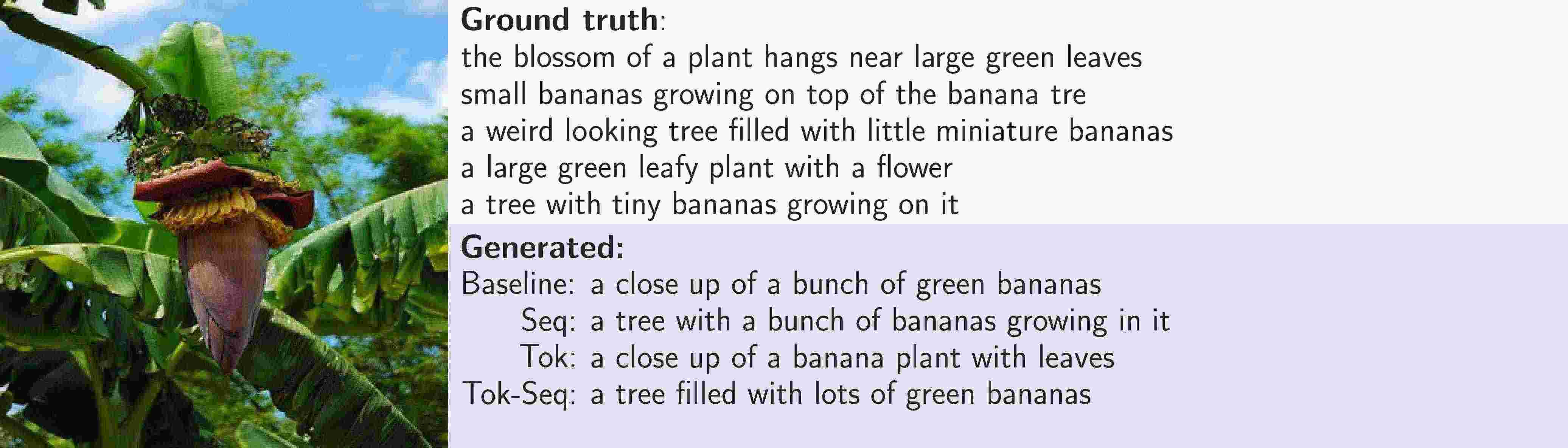} & \includegraphics[width=.49\textwidth]{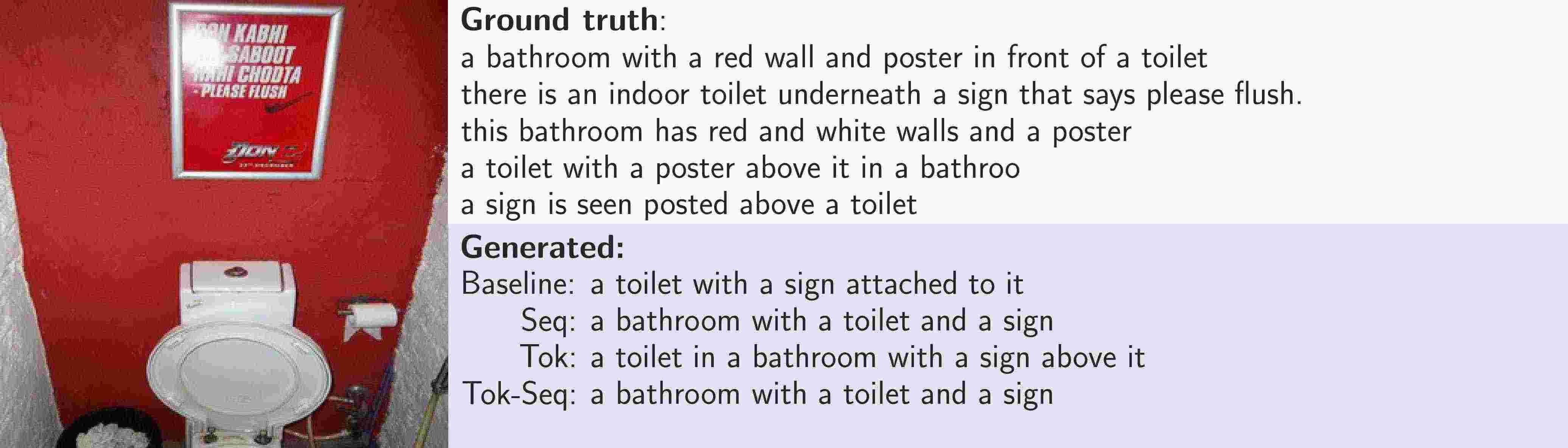} \\
\includegraphics[width=.49\textwidth]{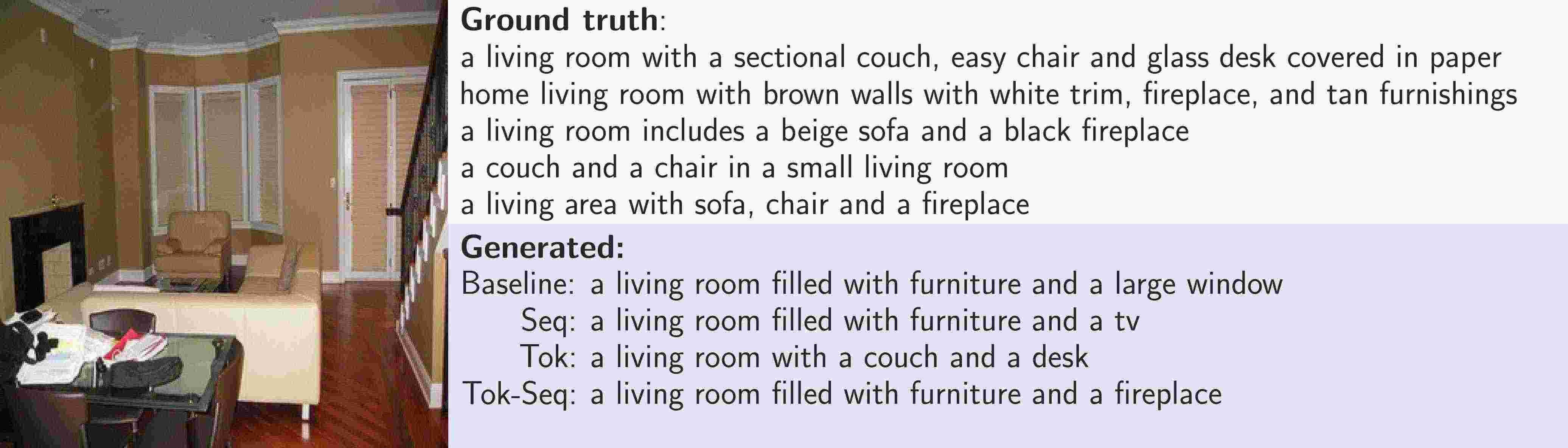} & \includegraphics[width=.49\textwidth]{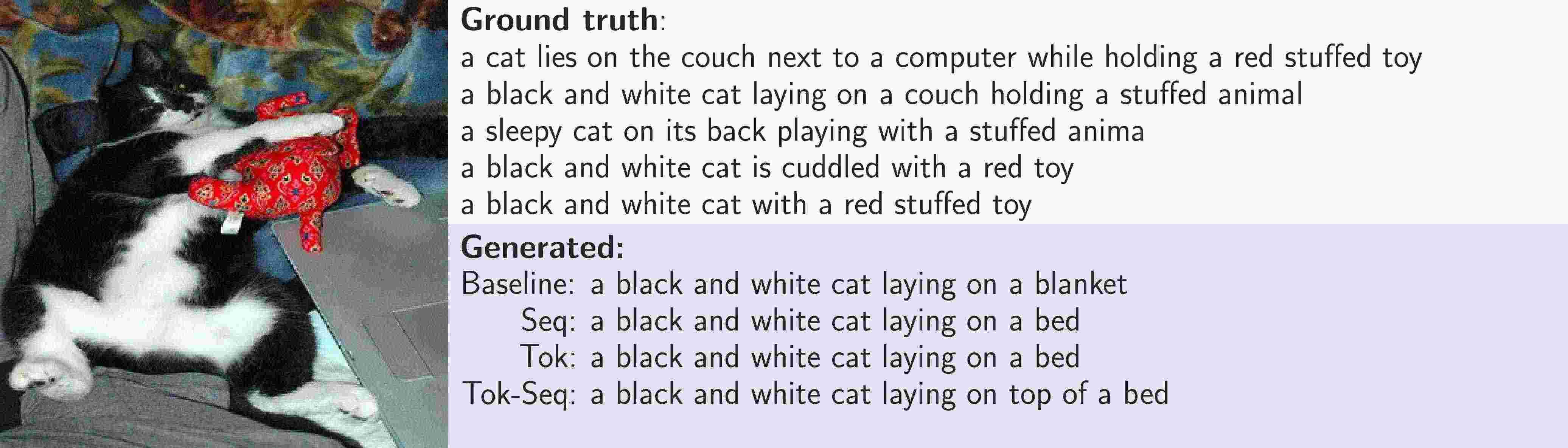} \\

\includegraphics[width=.49\textwidth]{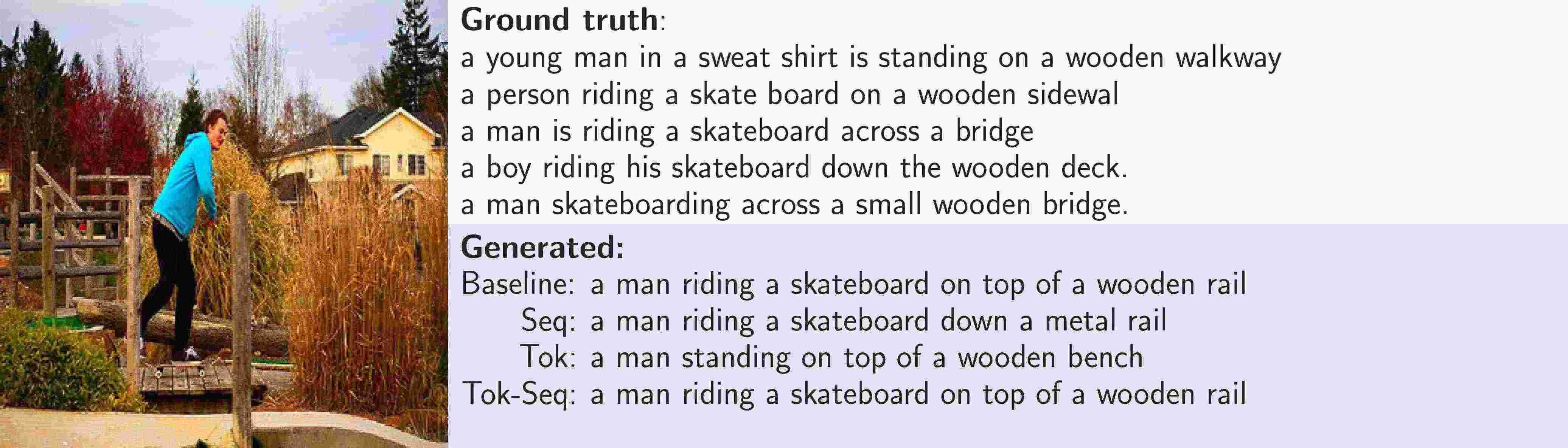} & \includegraphics[width=.49\textwidth]{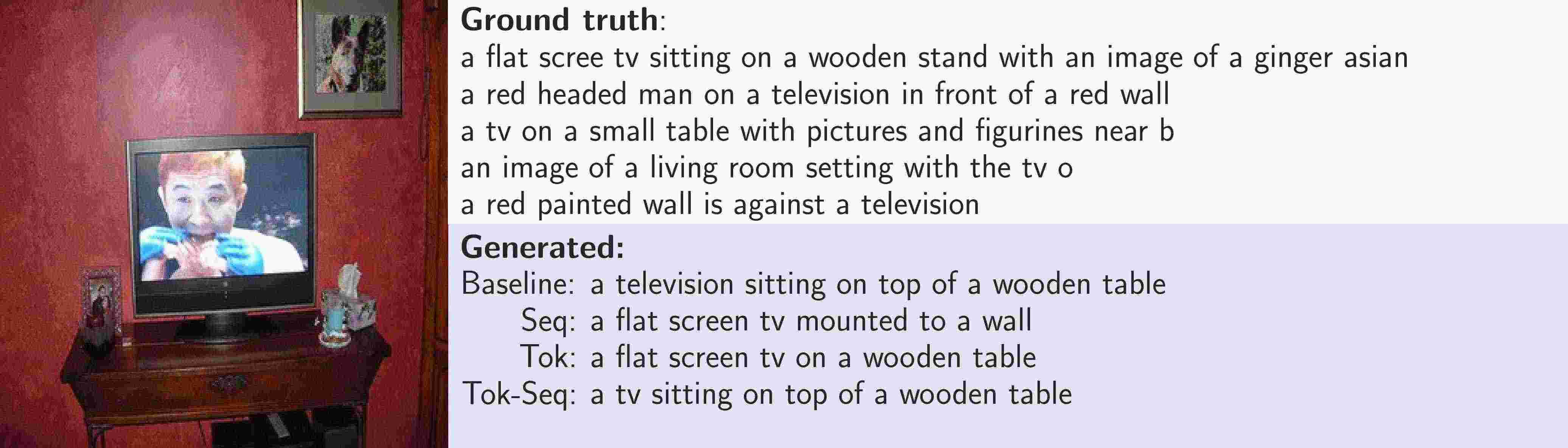} \\
\includegraphics[width=.49\textwidth]{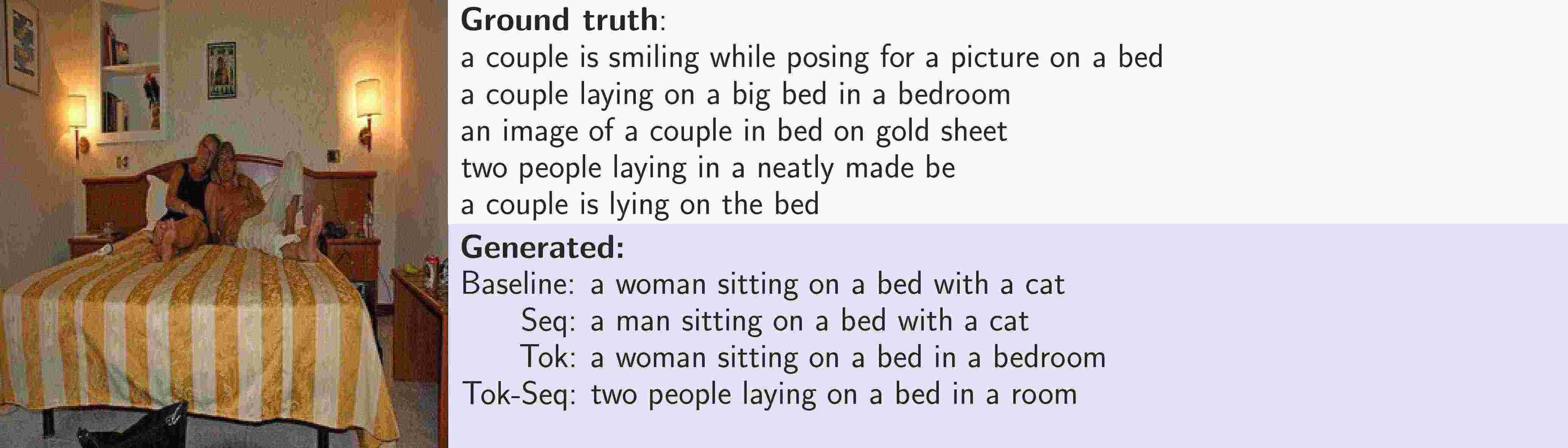} & \includegraphics[width=.49\textwidth]{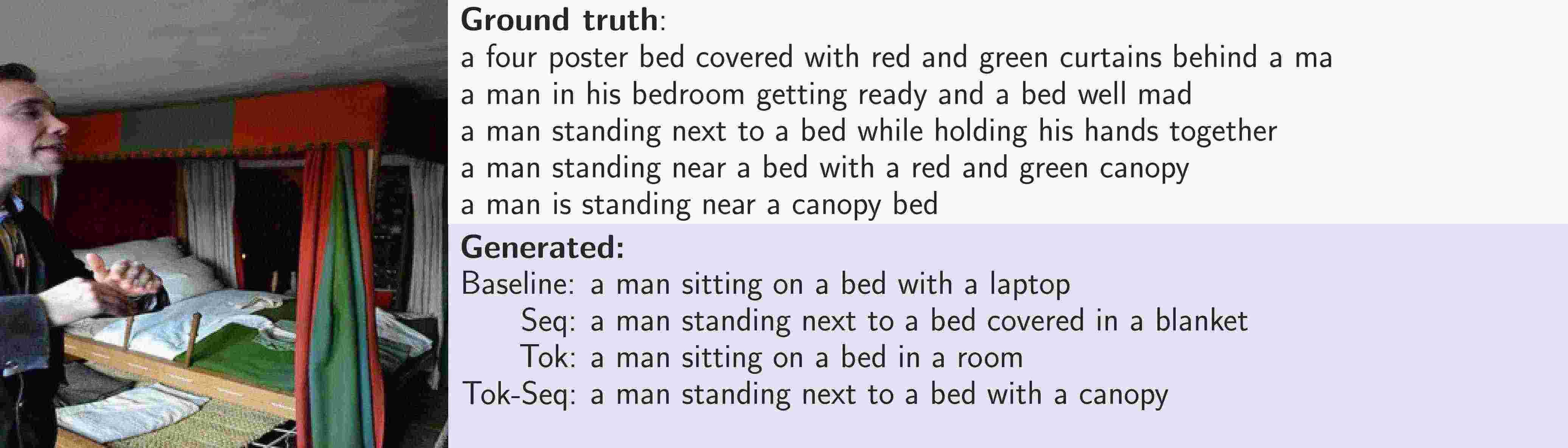} \\
\includegraphics[width=.49\textwidth]{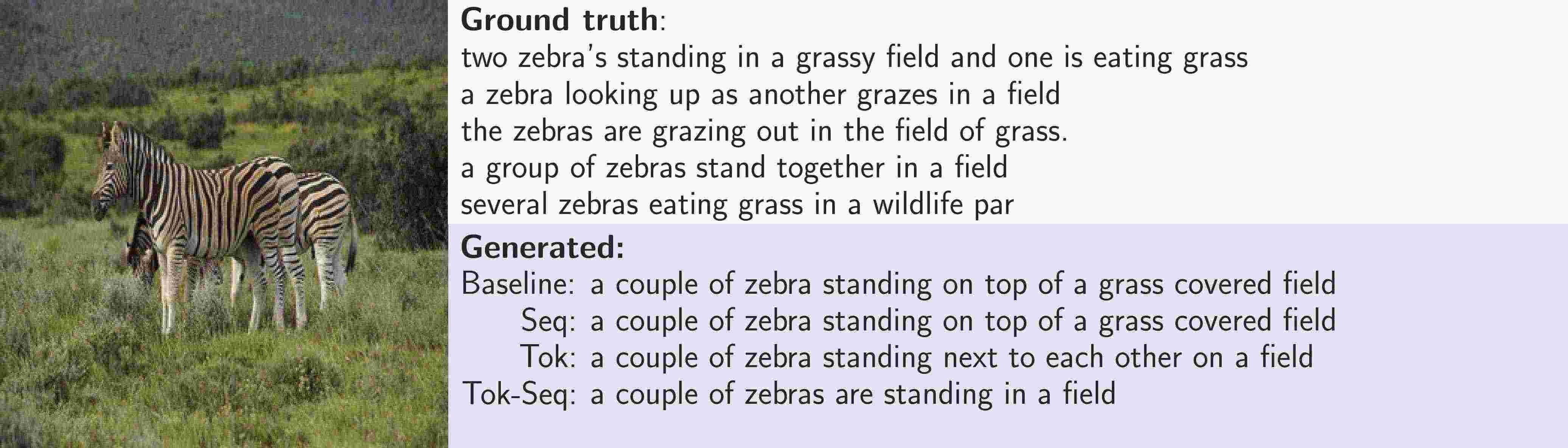} & \includegraphics[width=.49\textwidth]{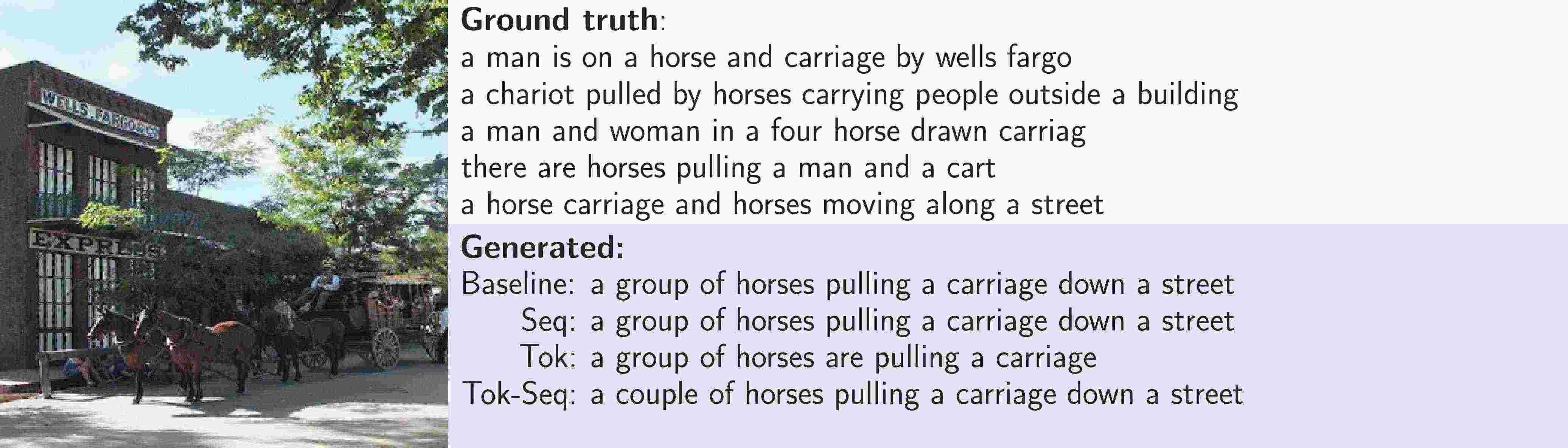} \\
\includegraphics[width=.49\textwidth]{caps/344955} & \includegraphics[width=.49\textwidth]{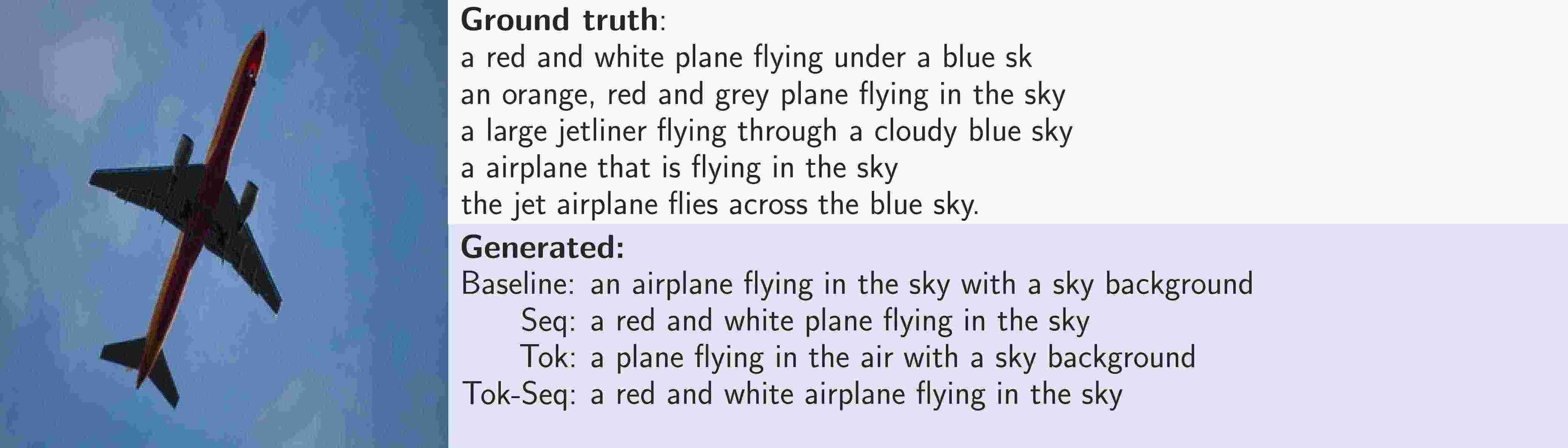} \\
\includegraphics[width=.49\textwidth]{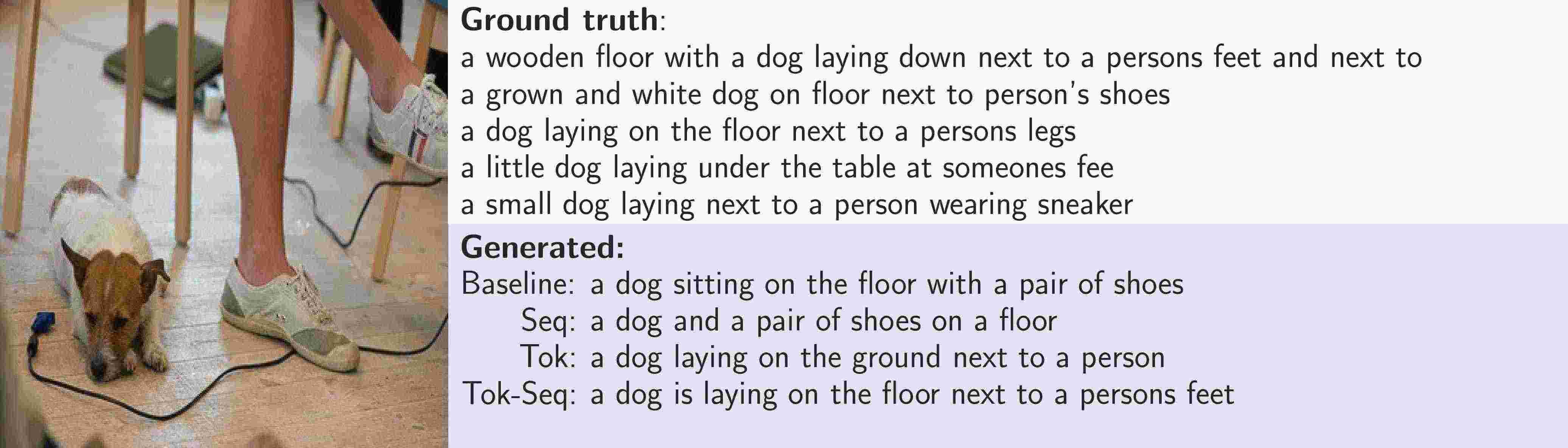} & \includegraphics[width=.49\textwidth]{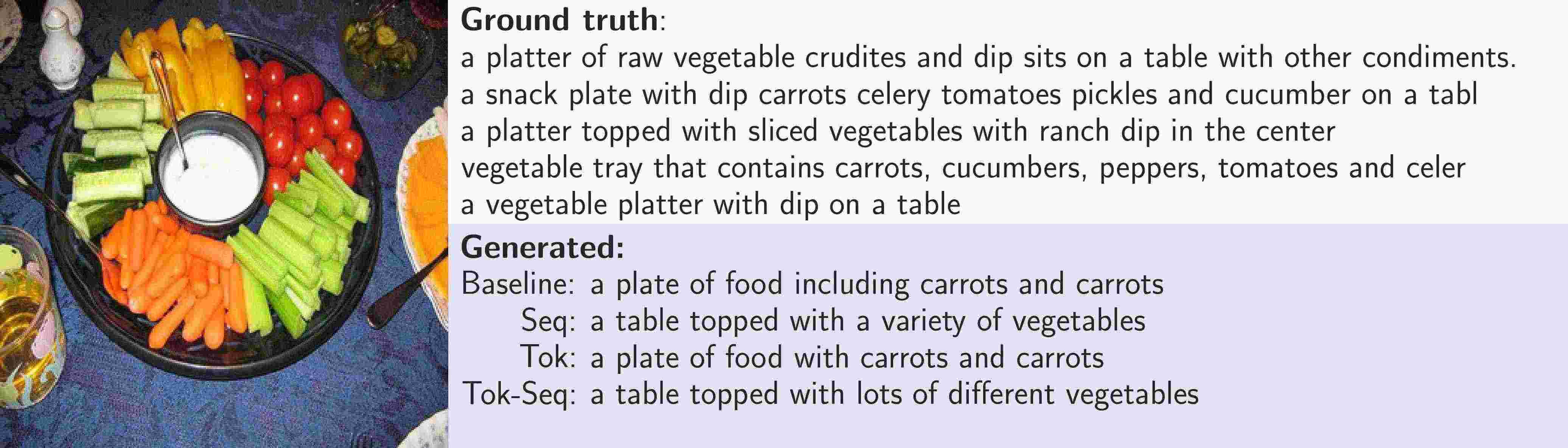} \\
\includegraphics[width=.49\textwidth]{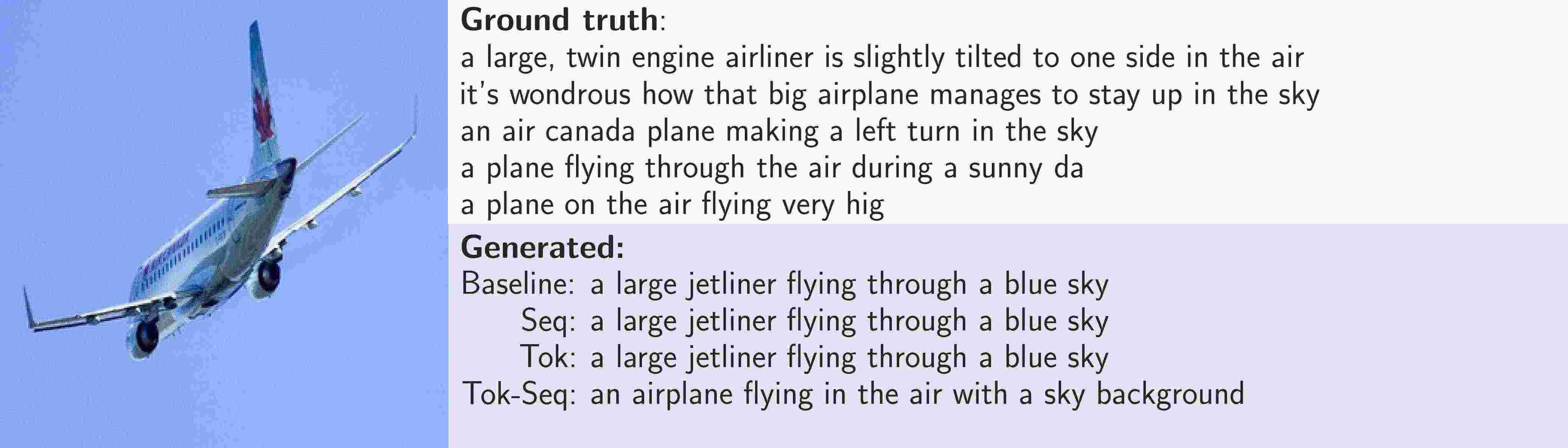}  & \includegraphics[width=.49\textwidth]{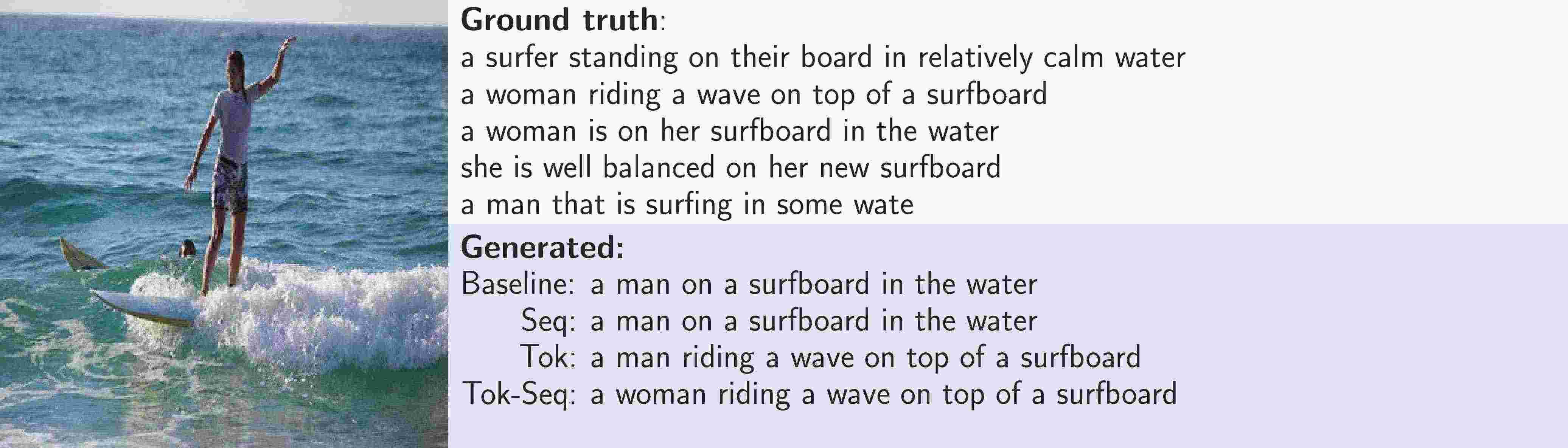} \\
\includegraphics[width=.49\textwidth]{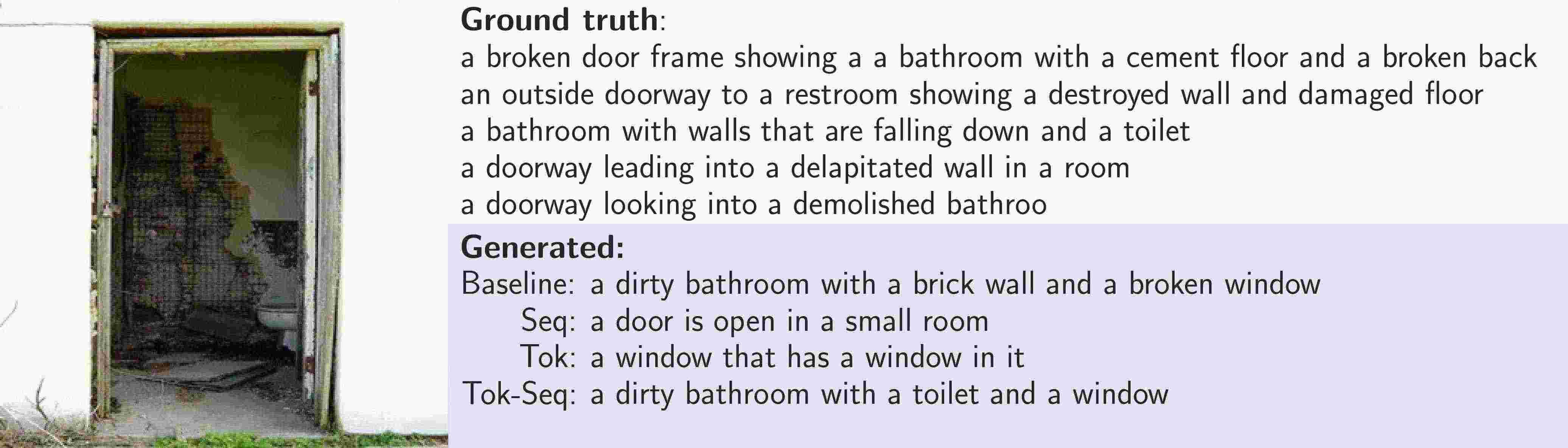}  & \includegraphics[width=.49\textwidth]{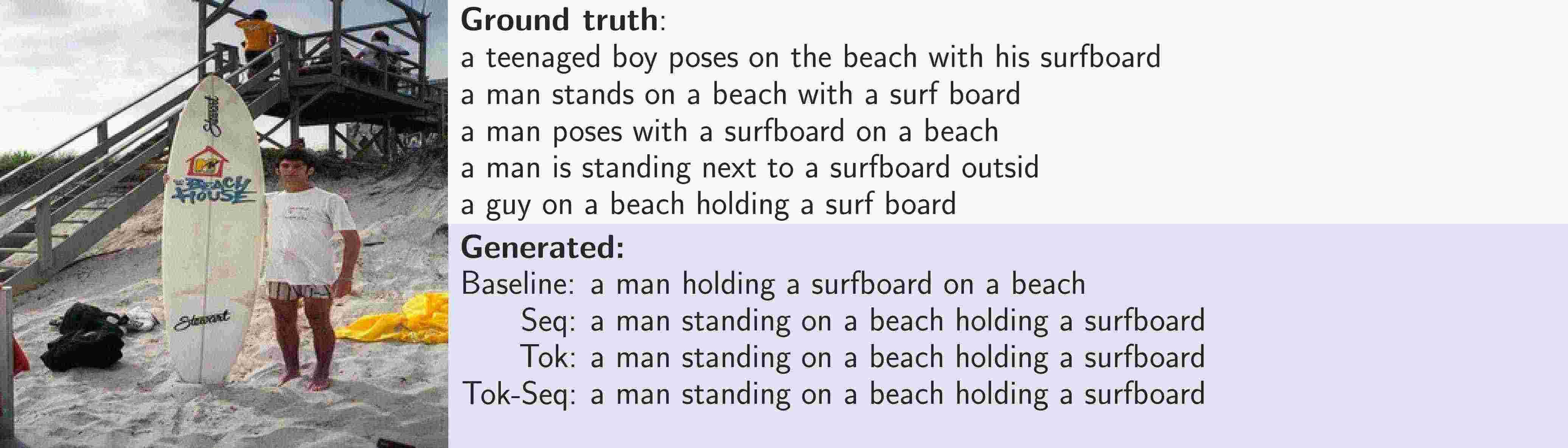} \\

\includegraphics[width=.49\textwidth]{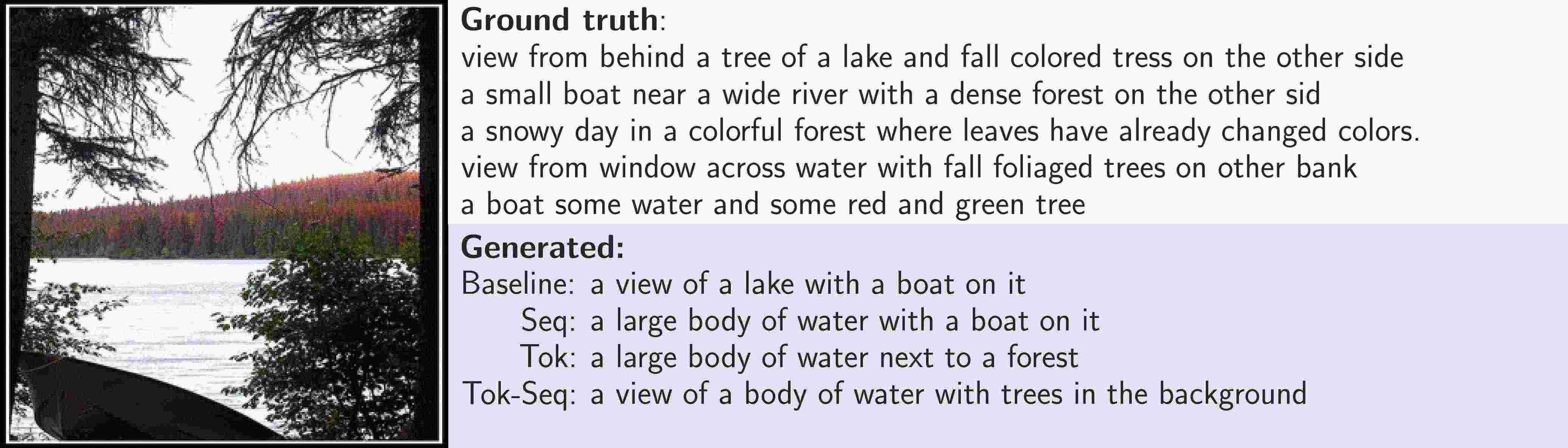}  & \includegraphics[width=.49\textwidth]{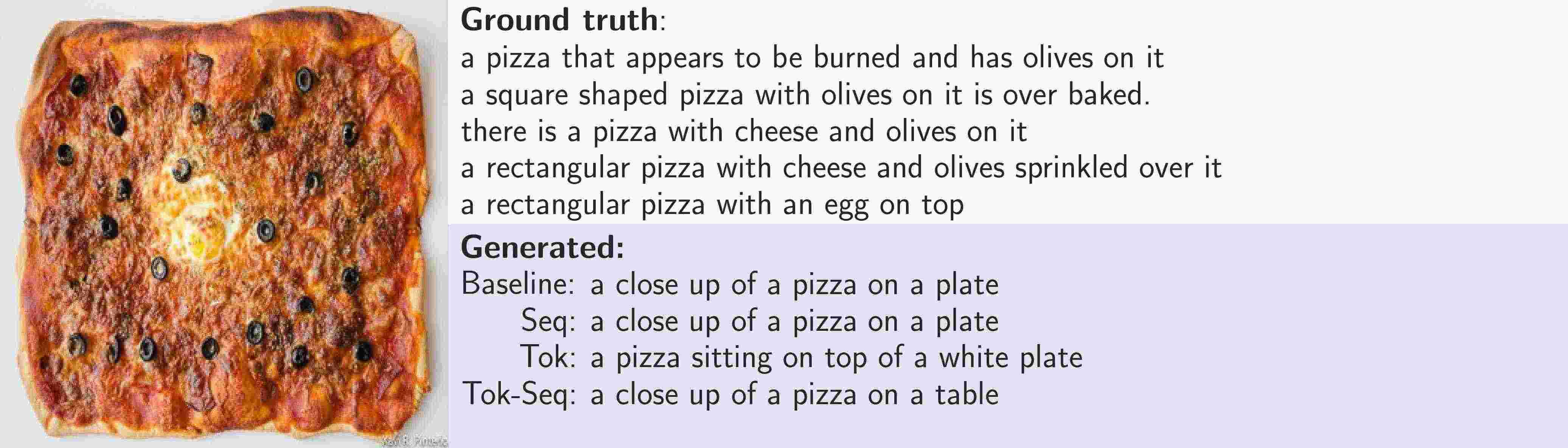} \\
\includegraphics[width=.49\textwidth]{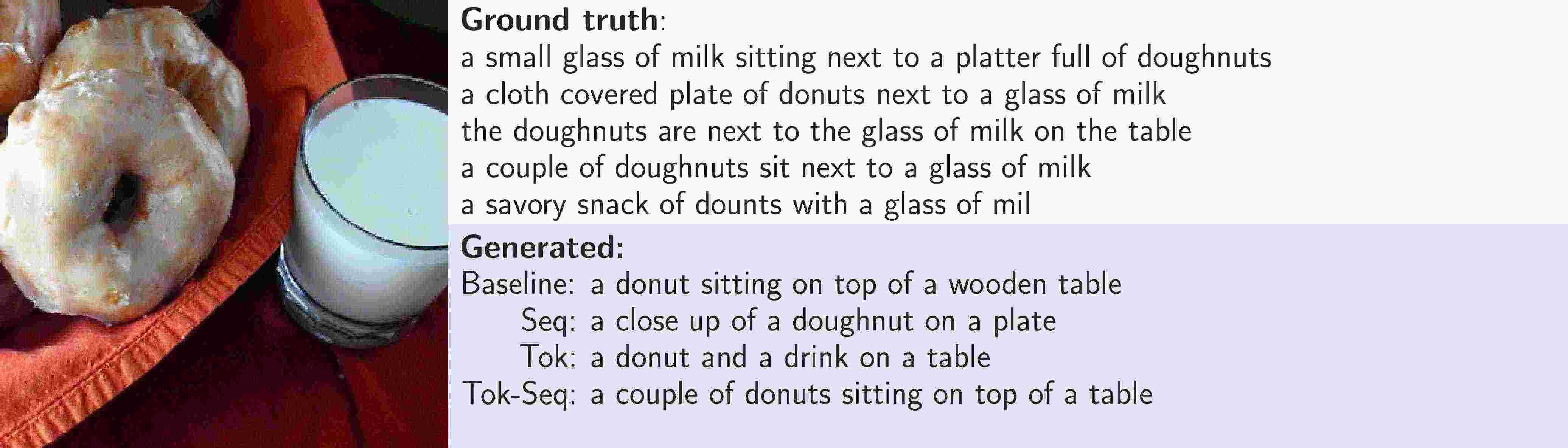}  & \includegraphics[width=.49\textwidth]{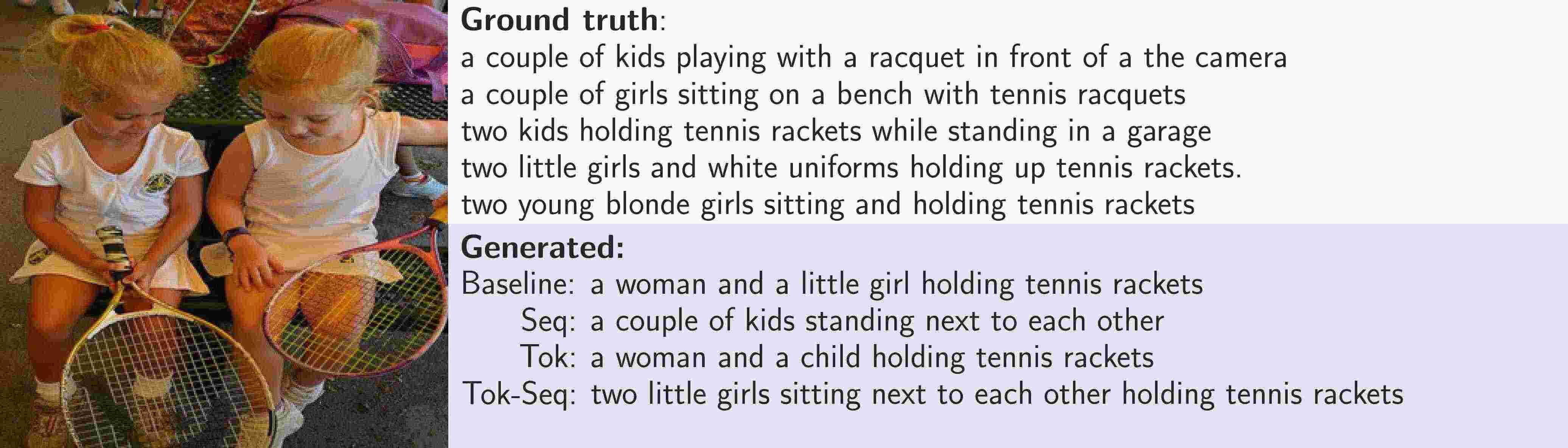} \\
\includegraphics[width=.49\textwidth]{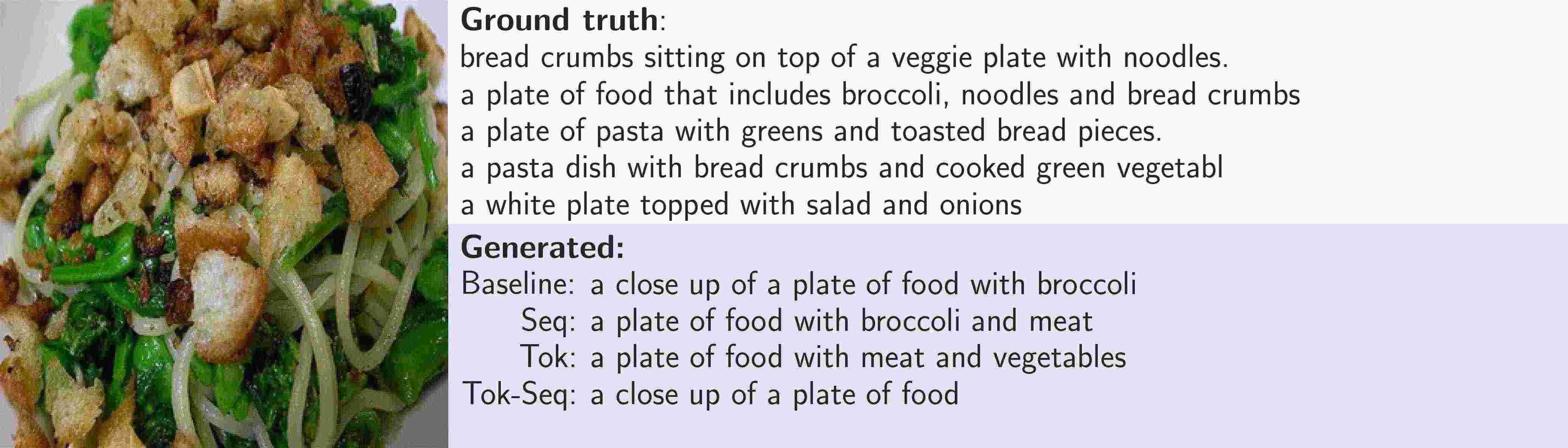}  & \includegraphics[width=.49\textwidth]{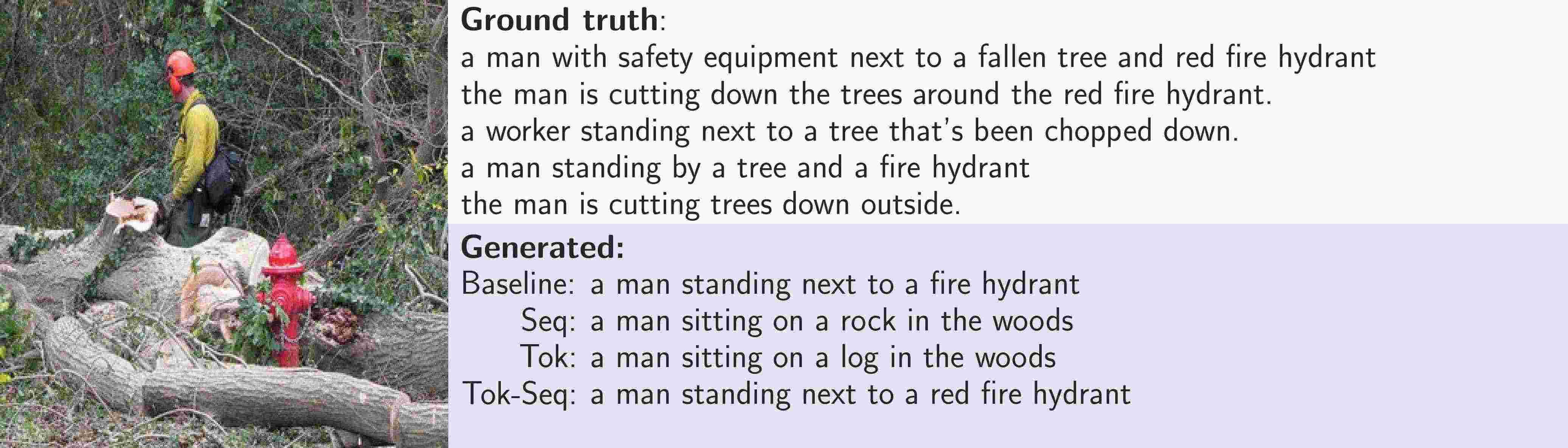} \\
\end{longtable}

\section{Neural machine translation}\label{appendix:nmt}
\subsection{Experimental setups}\label{appendix:nmtsetup}
\paragraph{WMT14 English-to-French}
We use the same experimental setting as \citet{bahdanau15iclr}: 12M paired sentences are used for training, 6,003 pairs for validation (news-test-2012 and news-test-2013)  and 3,003 test pairs (news-test-2014).
 After tokenization, 30k most frequent tokens are selected for the model's vocabulary.
We use an attentive encoder-decoder with a 2-layers bi-directionnal encoder of dimension $d=2000$ and a single-layer decoder of dimension $d=2000$ as well. 
We use batches of size 80 and train for 3 epochs with Adam \citep{kingma15iclr} starting with a learning rate of 2e-4. 
 To generate translations  we use beam search of size five.
\paragraph{IWSLT14 German-to-English}
We use the same settings as \citep{ranzato16iclr}; the training set consists of 153k sentence pairs and 7k pairs are assigned to the validation and test sets.
After tokenization and lower-casing, we remove sentences longer than 50 tokens.
The English vocabulary has 22,822 words while the German has 32,009 words.
We use an attentive encoder-decoder with a single bi-directionnal encoder and decoder of dimension $d=128$.
 To generate translations  we use beam search of size five.
We use batches of size 32 and train for 40 epochs with Adam \citep{kingma15iclr} starting with a learning rate of 1e-3. 
\subsection{Examples}
\begin{table}[!ht]
\begin{tabular}{lp{14cm}}
\toprule
Source (en) & I think it's conceivable that these data are used for mutual benefit .\\
Target (fr) & J'estime qu'il est concevable que ces données soient utilisées dans leur intérêt mutuel .\\
MLE & Je pense qu'il est possible que ces données soient utilisées à des fins réciproques .\\ 
Tok-Seq & Je pense qu'il est possible que ces données soient utilisées pour le bénéfice mutuel .\\
\midrule
Source (en) & However , given the ease with which their behaviour can be recorded , it will probably not be long before we understand why their tails sometimes go one way , sometimes the other .\\
Target (fr)  & Toutefois , étant donné la facilité avec laquelle leurs comportements peuvent être enregistrés , il ne faudra sûrement pas longtemps avant que nous comprenions pourquoi leur queue bouge parfois d'un côté et parfois de l'autre .\\ 
MLE & Cependant , compte tenu de la facilité avec laquelle on peut enregistrer leur comportement , il ne sera probablement pas temps de comprendre pourquoi leurs contemporains vont parfois une façon , parfois l'autre .\\
Tok-Seq  & Cependant , compte tenu de la facilité avec laquelle leur comportement peut être enregistré , il ne sera probablement pas long avant que nous ne comprenons la raison pour laquelle il arrive parfois que leurs agresseurs suivent un chemin , parfois l'autre .\\
\midrule
Source (en) & The public will be able to enjoy the technical prowess of young skaters , some of whom , like Hyeres' young star , Lorenzo Palumbo , have already taken part in top-notch competitions .\\
Target (fr) & Le public pourra admirer les prouesses techniques de jeunes qui , pour certains , fréquentent déjà les compétitions au plus haut niveau , à l'instar du jeune prodige hyérois Lorenzo Palumbo .\\
MLE & Le public sera en mesure de profiter des connaissances techniques des jeunes garçons , dont certains , à l'instar de la jeune star américaine , Lorenzo , ont déjà participé à des compétitions de compétition .\\
Tok-Seq & Le public sera en mesure de profiter de la finesse technique des jeunes musiciens , dont certains , comme la jeune star de l'entreprise , Lorenzo , ont déjà pris part à des compétitions de gymnastique .\\
\bottomrule
\end{tabular}
\caption{WMT'14 English-to-French examples}
\end{table}

\begin{table}[!ht]
\begin{tabular}{lp{14cm}}
\toprule
Source (de) &  sie repräsentieren teile der menschlichen vorstellungskraft , die in vergangene zeiten <UNK> . und für alle von uns , werden die träume dieser kinder , wie die träume unserer eigenen kinder teil der geographie der hoffnung .
\\
Target (en) &  they represent branches of the human imagination that go back to the dawn of time . and for all of us , the dreams of these children , like the dreams of our own children , become part of the naked geography of hope .
\\
MLE & they represent parts of the human imagination that were blogging in the past time , and for all of us , the dreams of these children will be like the dreams of our own children part of hope .\\ 
Tok-Seq & and they represent parts of the human imagination that live in past times , and for all of us , the dreams of these children , like the dreams of our own children are part of hope . \\
\midrule
Source (de) & und ja , vieles von dem , was heute gesagt wurde , berührt mich sehr , weil viele , viele schöne äußerungen dabei waren , die ich auch durchlebt habe .
 \\
Target (en) &  and yes , a lot of what is said today really moves me , because many , many nice statements were made , which i also was part of .
\\
MLE & and yes , a lot of what 's been said to me today , i got very , very , very , very beautiful expressions that i used to live through .\\ 
Tok-Seq & and yes , a lot of what 's been told today is very , very much , because many , lots of beautiful statements that i 've been through . \\ 
\midrule
Source (de) & noch besser , er wurde in <UNK> nach den angeblich höchsten standards der nachhaltigkeit gezüchtet . \\
Target (en) &  even better , it was <UNK> to the supposed highest standards of sustainability .  \\
MLE & even better , he has been bred in love with the highest highest standards of sustainability . \\ 
Tok-Seq & even better , he was raised in terms of dignity , the highest standards of sustainability .\\ 
\bottomrule
\end{tabular}
\caption{IWSLT'14 German-to-English examples}
\end{table}
\end{document}